\documentclass[sn-mathphys,Numbered]{sn-jnl}% Math and Physical Sciences Reference Style
\usepackage{graphicx}%
\usepackage{subcaption}
\usepackage{multirow}%
\usepackage{amsmath,amssymb,amsfonts}%
\usepackage{amsthm}%
\usepackage{mathrsfs}%
\usepackage[title]{appendix}%
\usepackage{xcolor}%
\usepackage{textcomp}%
\usepackage{manyfoot}%
\usepackage{booktabs}%
\usepackage{algorithm}%
\usepackage{algorithmicx}%
\usepackage{algpseudocode}%
\usepackage{listings}%
\usepackage{fixmath}
\usepackage{natbib}
\theoremstyle{thmstyleone}%
\theoremstyle{thmstyletwo}%

\theoremstyle{thmstylethree}%

\begin{document}

\title[Article Title]{WebQAmGaze: A Multilingual Webcam Eye-Tracking-While-Reading Dataset}


\author*[1]{\fnm{Tiago} \sur{Ribeiro}}\email{tiri@di.ku.dk}

\author[1]{\fnm{Stephanie} \sur{Brandl}}\email{brandl@di.ku.dk}

\author[1]{\fnm{Anders} \sur{Søgaard}}\email{soegaard@di.ku.dk}

\author[1,2]{\fnm{Nora} \sur{Hollenstein}}\email{nora.hollenstein@hum.ku.dk}

\affil[1]{\orgdiv{University of Copenhagen}, \country{Denmark}}
\affil[2]{\orgdiv{University of Zurich}, \country{Switzerland}}

\abstract{We present \textsc{WebQAmGaze}, a multilingual low-cost eye-tracking-while-reading dataset, designed as the first webcam-based eye-tracking corpus of reading to support the development of explainable computational language processing models. \textsc{WebQAmGaze} includes webcam eye-tracking data from 600 participants of a wide age range naturally reading English, German, Spanish, and Turkish texts. Each participant performs two reading tasks composed of five texts each, a normal reading and an information-seeking task, followed by a comprehension question. We compare the collected webcam data to high-quality eye-tracking recordings. The results show a moderate to strong correlation between the eye movement measures obtained with the webcam compared to those obtained with a commercial eye-tracking device.
When validating the data, we find that higher fixation duration on relevant text spans accurately indicates correctness when answering the corresponding questions.
This dataset advances webcam-based reading studies and opens avenues to low-cost and diverse data collection. \textsc{WebQAmGaze} is beneficial to learn about the cognitive processes behind question-answering and to apply these insights to computational models of language understanding.}




\maketitle

\section{Introduction}

Eye movement data is beneficial for natural language processing (NLP) models because it provides direct access to human language processing signals \citep{mishra2018cognitively}. Eye-tracking recordings can be leveraged to augment NLP models by providing a human inductive bias \citep{hollenstein-etal-2020-towards,barrett-etal-2018-sequence} or to evaluate and analyze the inner workings of the models and increase their explainability \citep{hollenstein-etal-2019-cognival,sood-etal-2020-interpreting,brandl-hollenstein-2022-every}. With the emergence of machine learning models in many fields of application, the need to explain the predictions of these models arises \citep{sogaard2021explainable}. More concretely, where readers look when searching for the answer to a question in a text can provide information to ensure transparency and explainability in computational models of question-answering.

Eye movement data recorded from natural reading has been used to improve models for various NLP tasks such as document summarizing tasks \citep{xu_user-oriented_2009}, part of speech tagging \citep{barrett_weakly_2016}, and named entity recognition \citep{hollenstein-zhang-2019-entity}, among others. The eye movements are translated into engineered reading time features, such as the number, location, and duration of fixations, which reflect the various stages of linguistic processing during language comprehension \citep{hollenstein-etal-2020-towards}. These features are then included as additional input to the machine learning models for NLP tasks. Reading time features have been shown to correlate to the attention mechanism in computational language processing models \citep{hollenstein-beinborn-2021-relative}. For this reason, there is a natural jump to attention models such as BERT, where the eye-tracking features can be used directly to weigh the attention layer \citep{mcguire2021sentiment,dong2022gazby,eberle-etal-2022-transformer} and improve the task performance of the model. 

However, these machine learning approaches rely on large text datasets and are thus often limited by the size and availability of existing eye-tracking datasets, which require expensive equipment and participants to be present in a lab to be collected. Dataset availability is also sparse, with most reading stimuli still being collected in English. It is also common for datasets to provide a specific reading task since different gaze patterns are elicited depending on the task participants are primed on, e.g., reading during linguistic annotation or information-seeking reading \citep{hollenstein-etal-2020-zuco,malmaud-etal-2020-bridging}. Therefore, it is difficult to transfer eye movement measures from specific datasets to other domains or tasks. Moreover, it opens the question of whether eye movement data from normal reading or task-specific reading is more beneficial for NLP models.

In light of these challenges, in this work, we present \textsc{WebQAmGaze}, a multilingual webcam eye-tracking-while-reading dataset tailored to be used not only for reading research and comparisons between low-quality and high-quality gaze recordings but also in machine learning-based NLP applications. To enable a large-scale experiment setup, the data is collected through the crowd-sourcing platforms \textit{Amazon Mechanical Turk} and \textit{Cognition}, paired with the open-source libraries \textit{jsPsych} \citep{de_leeuw_jspsych_2015} and \textit{WebGazer} \citep{papoutsaki_webgazer_2016}. To ensure the adequacy of the text stimuli for their use for downstream NLP applications, we select texts in multiple languages (English, Spanish, German, and Turkish) from an open-source question-answering dataset.
The \textsc{WebQAmGaze} data used and related experiment and analysis code are available online.\footnote{\url{https://github.com/tfnribeiro/WebQAmGaze}}

For the data collection, we employ two experiment paradigms, a \textit{normal reading} (NR) task, where participants read a continuous text and answer a comprehension question on the next screen, and an \textit{information-seeking} (IS) task, where participants are presented with the question they have to answer \textit{before} reading the text, followed by the text itself and the question again. Previous work has shown that information-seeking reading results in faster reading speed and higher omission rates, i.e., fewer words are fixated during such a search task when compared to normal reading \citep{hollenstein-etal-2020-zuco}. This type of reading also leads to lower alignment with NLP models \citep{eberle-etal-2022-transformer}. By collecting eye-tracking data from both tasks, we provide the possibility to analyze this behavior in webcam recordings. We hypothesize that a higher number of fixations on the relevant target spans in the text will result in participants answering to the questions correctly.

\section{Related Work}

In this section, we discuss previous work in this area of research. We focus on eye-tracking-while-reading studies (Section \ref{sec:relw-et-reading}), the usage of eye-tracking data in NLP models (Section \ref{sec:relw-et-nlp}), and recent progress in webcam-based eye-tracking (Section \ref{sec:relw-webcam-et}), especially with respect to reading research (Section \ref{sec:relw-webcam-reading}).

\subsection{Eye-Tracking-While-Reading}\label{sec:relw-et-reading}

When looking at eye-tracking data collected in reading tasks, one can observe that readers move their eyes rapidly across the text and fixate on different words, often skipping words entirely, other times fixating them for longer and sometimes even returning to previous words \citep{rayner2012psychology,liversedge2000saccadic}. These patterns are linked with cognitive-linguistic processes and reveal insights not only into the reader's comprehension of the text but also highlight linguistic properties of the words that are fixated, such as word frequency and word class \citep{rayner1998eye,radach2004theoretical}. \\
\indent Eye-tracking technology has been used for years to study reading behavior and language comprehension \citep{winke2013eye}. The availability of eye-tracking-while-reading corpora is increasing steadily, with high-quality datasets covering more and more languages (e.g., \citet{hollenstein2018zuco,cop2017presenting,berzak2022celer,siegelman_expanding_2022_meco}).
Research is moving more towards ecologically valid experiments of reading by including naturally occurring texts and allowing participants to read at their own speed (e.g., \citet{desai2016toward,kuperman2023text,hollenstein-etal-2022-copenhagen}). While high-quality lab-controlled eye-tracking studies have well-defined standards for testing, reporting, and analysis \citep{dunn2023minimal,holmqvist2023eye}, the territories of using webcam-based eye-tracking for reading studies remain relatively uncharted.

\subsection{Leveraging Eye Movement Data for Natural Language Processing}\label{sec:relw-et-nlp}

Eye movement data of reading provides rich insights into cognitive processes of language understanding. The signals can be used to modulate the inductive bias of machine learning models towards
more cognitively plausible processing which can
increase model performance \citep{hollenstein2019advancing,beinborn2023cognitive}.
For these reasons, a variety of datasets have been created to study different properties \citep{mathias2021survey}, such as the \textit{ZuCo} \citep{hollenstein2018zuco}, \textit{GECO} \citep{cop2017presenting} or \textit{PROVO} \citep{luke_provo_2018} corpora, each introducing different reading tasks and texts, which usually are decided based on the research question being answered. Tasks can range from self-paced reading of novels to existing NLP task-specific corpora, such as sentiment banks or relation detection datasets, where patterns in eye-tracking data highlight certain linguistic processing patterns. 

In natural language processing, eye-tracking data has been successfully used in tasks such as part-of-speech tagging \citep{barrett_weakly_2016}, readability assessment \citep{gonzalez-garduno_using_2017}, sentiment analysis \citep{mishra-etal-2016-leveraging, barrett_sequence_2018}, named entity recognition (NER) \citep{hollenstein-zhang-2019-entity}, among others, and they all report improvements over baselines without eye-tracking features \citep{mathias2021survey}. Additionally, eye movement data have been shown to correlate to the attention mechanisms in computational language processing models \citep{hollenstein-beinborn-2021-relative,eberle-etal-2022-transformer,morger-etal-2022-cross}.

Nevertheless, researchers highlight the importance of high-quality data for capturing human gaze interaction \citep{holmqvist2012eye} and for using eye-tracking features in an NLP context to ensure the validity of the results \citep{hollenstein-etal-2020-towards}. This leaves the open question of how viable gaze data from low-cost methods can be for NLP applications, where lower quality can be attributed to lower sampling rates, calibration accuracy, and eye-tracker precision.
Moreover, it is still unclear what the best approach to employ the extracted features is (e.g., as attention proxies or input features to the neural network models), and to which granularity (e.g., word-level or sentence-level features) they should be used to obtain the best results given a task.

In short, there has been increasing evidence of eye-tracking being useful for solving NLP tasks, but the use of low-cost datasets is yet to be explored. While it presents methodological challenges, the possibility of collecting large amounts of low-cost eye-tracking data can enable coverage of more diverse reader populations, which can prove important to better understand phenomena, such as attention while reading. We follow the strategy proposed by \citet{andersson2010sampling}, which shows how low sampling frequency can be compensated with more data processing and post-experiment adjustments of measures.

It is yet to be investigated what impact such low-quality eye-tracking data will have on downstream NLP tasks such as question answering. In this project, we create a resource that seeks to provide an initial look at how low-cost eye-tracking data can be used to extract word-level fixation measures and provide a large enough sample size to allow researchers to evaluate its usefulness for NLP models.

\subsection{Webcam-Based Eye-Tracking}\label{sec:relw-webcam-et}

Low-cost video-based eye-tracking has been investigated since the last decade \citep{wei_low_2016,papoutsaki_scalable_2015}. Video-based tracking is implemented with appearance-based gaze-estimation models. Unlike model-based approaches, appearance-based models focus on extracting eye movement measures based on the natural appearance of the eye from a regular camera \citep{sugano2013appearance}. These regression models are often developed based on datasets, in which participants are asked to fixate specific regions of the screen. 
However, these methods require a large number of calibration points, which is impractical when performing online experiments and head movements remain the main challenge for these models \citep{lu_adaptive_2014}. 
%The models use both the image of the participants as well as image saliency \citep{sugano2013appearance}. %\citet{lu_adaptive_2014} describe an adaptive linear regression method able to obtain an error as low as 0.97º, starting from 9 calibration points. }

Publicly available libraries such as \textit{TurkerGaze} \citep{xu_turkergaze_2015} and \textit{WebGazer} \citep{papoutsaki_webgazer_2016} are implemented in JavaScript, which allows the program to run independently on any platform and to be easily incorporated into a webpage. \textit{TurkerGaze}, for instance, aimed to explore webcam eye-tracking using \textit{Amazon Mechanical Turk} to determine the saliency of objects in images and showed promising results. To tackle the issue of head movements, \textit{TurkerGaze} utilizes multiple calibration and validation steps between tasks, while \textit{WebGazer} allows for calibration based on user inputs.

More recent approaches include \textit{SearchGazer} \citep{papoutsaki_searchgazer_2017}, which was developed for information retrieval and does not require any calibration, as it automatically calibrates itself by using clicks from users. \textit{SearchGazer} has been used for personalized text summarization, where gaze point density was used as a metric for sentence saliency \citep{dubey_wikigaze_2020}. These approaches, however, focus on extracting heatmaps, which are then used as features for a model to help reduce the search space. 
Finally, researchers are currently evaluating the quality of affordable eye-tracking on smartphones \citep{valliappan2020accelerating}. 
These advances show promise in the technology with models achieving better performance than the base models lacking these features.

\textit{WebGazer} \citep{papoutsaki_webgazer_2016} is our chosen library for this work. While it is similar to \textit{TurkerGaze}, it focuses on providing real-time gaze prediction and is actively maintained.

\subsection{Webcam-Based Eye-Tracking-While-Reading}\label{sec:relw-webcam-reading}

When it comes to webcam-based eye-tracking during reading, less research is available as of yet. A recent study performs a direct comparison of data collected from a commercial eye-tracker to that of webcam-based eye-tracking in a reading task and an accuracy task \citep{lin_eye_2022}. For the latter, results showed that the webcam-based eye-tracking algorithm distinguishes movements on the horizontal axis (right/left) more clearly, as opposed to the vertical axis. This is a challenge to overcome in reading experiments, especially if lines of text are close together. Nevertheless, the preliminary results are encouraging showing that differences between age groups can be identified by the webcam setup. \citet{guan_analysis_2022} use \textit{WebGazer} to perform a study on L2 English readers ($n=32$). They investigate reading comprehension based on engineered features, such as fixation counts on pages and lines, and regressions (going to a previous page). Their results show that these features are indicative of participants responding correctly. 
Lastly, \citet{hutt2023webcam} also leverage \textit{WebGazer} due to its benefits for online educational technologies, to detect cognitive states of mind wandering during reading comprehension. This webcam-based eye-tracking data proved to be sufficiently accurate and precise to predict both task-unrelated thought and reading comprehension. 

To the best of our knowledge, there is yet to be an extensive study of webcam eye-tracking in a natural reading setting, in order to extract linguistic features on word-level. For this reason, we contribute to this line of research by comparing our results to those of datasets compiled using lab-based eye-tracking equipment, such as MECO \citep{siegelman_expanding_2022_meco}, and provide recommendations on how to collect data in such a remote setting. We also use a methodology to evaluate differences in reading behavior in different tasks similar to that employed in \citet{malmaud-etal-2020-bridging}, which we will describe in the following sections.

\section{The \textsc{WebQAmGaze} Dataset}

We describe the process of compiling \textsc{WebQAmGaze}, a data resource that provides both raw eye movement recordings as well as pre-processed gaze fixations across 600 participants reading texts in 4 different languages: English (EN), German (DE), Spanish (ES), and Turkish (TR). We outline the experiment design (Section \ref{sec:exp-design}), the reading materials and their inclusion criteria (Section \ref{sec:read-materials}), as well as the chosen crowd-sourcing settings (Section \ref{sec:crowd-sourcing}). Thereafter, we report the details of the experiment protocol (Section \ref{sec:exp-structure}) and the presentation settings (Section \ref{sec:exp-presentation}).

The experiment design for this data collection was approved by the Ethics Commission of the Faculty of Humanities of the University of Copenhagen. Wherever possible, we follow the recommended empirical guidelines for reporting eye-tracking experiments \cite{holmqvist2023eye,dunn2023minimal}. However, we note that for webcam-based eye-tracking experiments, the experimental characteristics include higher variability than lab-controlled experiments (for instance, a wide range of screen sizes and sampling rates rather than one specific tracking device with a unique specified sampling rate).

\begin{table*}[t]
\centering
\caption{Reading materials included in the \textsc{WebQAmGaze} dataset, including the original datasets of the texts, the number of individual texts, the text length as the number of tokens (min, max, mean), and the mean sentence length as the average number of tokens per sentence.}\label{tab:dataset-stat}
\begin{tabular}{l|l|r|rrr|r}
\toprule
\textbf{Language} & \textbf{Origin Dataset} & \multicolumn{1}{l|}{\textbf{Texts}} & \multicolumn{3}{c|}{\textbf{Tokens}} & \multicolumn{1}{c}{\textbf{Length}} \\ \midrule
\multirow{2}{*}{English} & MECO  & 4  & \multicolumn{1}{r|}{184} & \multicolumn{1}{r|}{218} & 203.5 & 24.7 \\
                         & XQuAD & 97 & \multicolumn{1}{r|}{31}  & \multicolumn{1}{r|}{130} & 97.2  & 32.6 \\ \hline
\multirow{2}{*}{German}  & MECO  & 2  & \multicolumn{1}{r|}{178} & \multicolumn{1}{r|}{192} & 185.0 & 20.7 \\
                         & XQuAD & 45 & \multicolumn{1}{r|}{26}  & \multicolumn{1}{r|}{115} & 82.0  & 29.1 \\ \hline
\multirow{2}{*}{Spanish} & MECO  & 1  & \multicolumn{1}{r|}{195} & \multicolumn{1}{r|}{195} & 195.0 & 24.4 \\
                         & XQuAD & 64 & \multicolumn{1}{r|}{35}  & \multicolumn{1}{r|}{131} & 98.7  & 34.5 \\ \hline
\multirow{2}{*}{Turkish} & MECO  & 2  & \multicolumn{1}{r|}{151} & \multicolumn{1}{r|}{157} & 155.0 & 19.4 \\
                         & XQuAD & 36 & \multicolumn{1}{r|}{49}  & \multicolumn{1}{r|}{105} & 77.2  & 21.7 \\ \hline    
                         \end{tabular}
\end{table*}

\subsection{Experiment Design}\label{sec:exp-design}

The \textsc{WebQAmGaze} dataset is divided into two types of reading paradigms following the approach in \citet{malmaud-etal-2020-bridging}: A \textit{normal reading} task (NR), and an \textit{information-seeking reading} task (IS). There is no time limit for reading the texts in either scenario.

In the NR task, the participants are instructed to read a text carefully at their own speed and to press the spacebar to proceed to the next screen where they are asked a comprehension question about the text. The question is either a true/false question (for the MECO texts) or an open-answer question (for the XQuAD texts).

In the IS task, the participants are first presented with the question they will need to find an answer to in the text. The text is then presented on the next screen and the participants are instructed to press the spacebar as soon as they find the answer in the text. Then they are shown the question once again and they have to type their answer into a free text field. 

\subsection{Reading Materials}\label{sec:read-materials}
For the construction of the \textsc{WebQAmGaze} dataset, we use the freely available texts from the multilingual XQuAD dataset \citep{artetxe-etal-2020-cross} that can be used to test machine text comprehension models, meaning that the texts are accompanied by human annotations for question-answering.

XQuaD is a subset of the SQuAD question-answering dataset \citep{rajpurkar-etal-2016-squad} that has been translated into other languages. It contains pairs of texts and questions, annotated with target spans and correct answers. We include these texts as they already include marked relevant spans containing the crucial information required to answer the questions. This is comparable to a human rationale and allows us to compare the annotated target span to the eye-tracking data we collect to investigate if the gaze information also reflects the annotated spans when the questions are responded to correctly. The data we collect can be used in the existing state-of-the-art approaches for explainable NLP and compared directly to investigate how beneficial the eye movement data is in reflecting human rationales.% It also spans multiple languages, which we believe is important to also investigate how different languages might result in different patterns. 

We also include texts from the MECO corpus \citep{siegelman_expanding_2022_meco}. This dataset contains NR eye-tracking data coupled with reading comprehension questions for 13 different languages. Including texts from this dataset allows us to compare the webcam data to those collected by a commercial eye tracker and to draw conclusions about the quality of webcam eye-tracking. We describe this comparison analysis in Section \ref{sec:meco-comparison}. 

We extract data for languages in which the two datasets (XQuaD and MECO) overlap: English, Spanish, German, Turkish, Greek, and Russian. For the \textsc{WebQAmGaze} dataset, we focus on the first four languages, leaving Greek and Russian for potential extensions in future work.

\subsubsection{Text Selection Criteria}

We include texts from both datasets according to the following criteria:

\paragraph{XQuAD Texts} We select texts from the XQuAD question-answering dataset that are at most 650 characters long. We do this to allow for fitting the texts into smaller screens and to avoid the experiment taking too much time to complete.

\paragraph{MECO Texts} The texts in the MECO corpus are significantly longer than those present in XQuAD. Therefore, to include them in the experiment, we extend the total character limit to 1300. This results in the MECO texts having a slightly smaller font and spacing compared to the texts in XQuAD and only a subset of the original MECO texts is included in \textsc{WebQAmGaze}.\\

\noindent We report the statistics along with selected linguistic properties of all texts included in the \textsc{WebQAmGaze} dataset in Table \ref{tab:dataset-stat}. The number of tokens (i.e., instances of a sequence of characters in some particular document that are grouped together as a useful semantic unit for processing \citep{schutze2008introduction}) and sentence lengths are obtained by using \textit{spaCy}'s\footnote{\url{https://spacy.io/api/tokenizer/}} tokenizer for all languages except Turkish, where we use \emph{TrTokenizer}.\footnote{\url{https://github.com/apdullahyayik/TrTokenizer}}

\begin{figure*}[t]
    \centering
    \includegraphics[width=\textwidth]{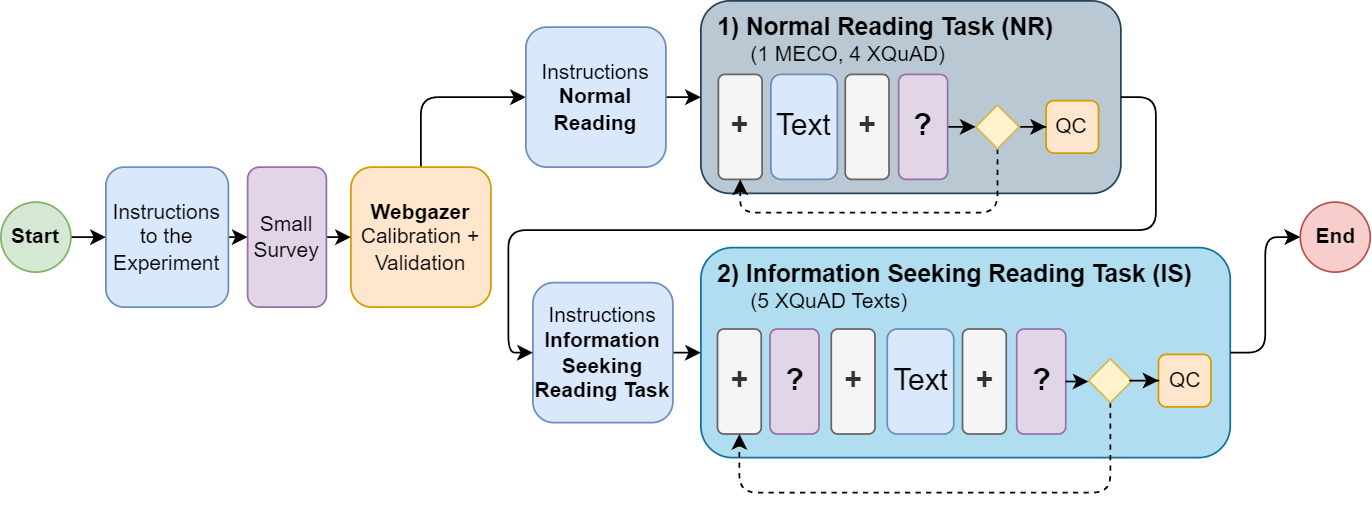}
    \caption{Experiment structure for the \textsc{WebQAmGaze} data collection. Light blue boxes represent reading pages, purple indicates input from the participants, orange indicates \textit{WebGazer} calibration and validation steps, and white boxes represent the screens with fixation crosses. Every second trial there is a quick calibration step, indicated by the yellow and orange boxes within the two reading tasks, i.e., there is a calibration after the 2\textsuperscript{nd} and 4\textsuperscript{th} trial in the NR task and the 1\textsuperscript{st} and 3\textsuperscript{rd} trial in the IS task.}
    \label{fig:experiment-procedure}
\end{figure*}

\subsection{Experiment Structure}\label{sec:exp-structure}

Eye-tracking data was collected using \textit{WebGazer}, an algorithm for webcam-based eye-tracking provided by \citet{papoutsaki_webgazer_2016}, who report an average visual angle error of $4.17^\circ$ in their initial publication.

The experiment follows the procedure illustrated in Figure \ref{fig:experiment-procedure}. Before the participants start with the experiment, they are asked to accept a consent form informing them of the requirements of the experiment. 
The participants are introduced to the two reading paradigms and are asked to fill out a survey of demographic information including age, native language, and fluency in the set's language. They are then instructed to close any unnecessary applications and are shown how to set up their screen to allow the best lighting conditions for \textit{WebGazer}. 
On the next screen, \textit{WebGazer} asks permission to access the webcam and start collecting data. Further instructions are provided on how to use the eye-tracker via a graphic with the following instructions: (1) Make sure your face is centered and nothing is obstructing the camera. (2) Do not move or tilt your head. Use only your eye movements to perform the task. (3) Do not sit too far or too close to the screen. (4) The image cannot be too dark. The images can be found in Appendix \ref{app:instructions}.

The participants are then prompted to set their browser to full-screen to allow for better calibration of \textit{WebGazer}. They are asked to keep their browsers in this mode until the end of the experiment. This ensures that the calibration grid matches the resolution of the participants' screen. The dimensions of the calibration grids are automatically adapted and mapped according to the screen resolution of the participant. The grids scale so that the points correspond to the same percentage in screens of different sizes. 
A 9-point calibration step is followed by a 5-point validation step. We reduce the validation and quick re-calibration steps to 5 points to reduce the total experiment duration and, in turn, the participant drop-out rate. 
Figures \ref{fig:calibration-grid} and \ref{fig:calibration-grid-quick}in Appendix \ref{app:calibration-grid} show the calibration grid used in the experiment. Additionally, examples of the validation results can be found in Appendix \ref{app:validation}.

The calibration accuracy is calculated as the percentage of gaze samples falling inside a $100px$ radius of the AOI of each point, and if the average calibration accuracy is lower than $60\%$ we ask participants to repeat the calibration step once more. If the average calibration accuracy after the second validation remains $<60\%$, we allow the participants to proceed to the reading tasks to reduce the participant drop-out rate. The reported calibration accuracy is the last average obtained from validation. 

The participants start with the NR task, where they first read instructions and then complete 5 texts (1 from MECO and 4 from XQuAD). This is followed by the IS task, where first the instructions and then 5 texts (all 5 from XQuAD) are shown. These 10 texts denote what we call a set $s$, which is a specific sequence of texts in a fixed order. In each set, 5 NR texts are followed by 5 IS texts. Each set is read by at least 9 participants.
These sets are created by randomly sampling from an available pool of texts until all texts have been used once in a trial, either NR or IS. After all texts have been used, new sets are created by selecting texts randomly, always including 1 MECO and 9 XQuAD texts. We perform the NR task first, as it is the most cognitively demanding task, followed by the less demanding IS task. 
A quick calibration (QC) step in the form of a 5-point calibration is done every 2\textsuperscript{nd} trial. This means a QC is performed after the 2\textsuperscript{nd} and 4\textsuperscript{th} trial in the NR task and the 1st and 3rd trials in the IS task. The experiment then terminates by quitting full-screen mode and asking participants to submit the HIT on the \textit{Amazon Mechanical Turk} platform. For each $s$ we collect data from at least 9 participants.

\subsection{Stimulus Presentation Settings}\label{sec:exp-presentation}

The stimuli are constructed using HTML and CSS. The CSS differs depending on the dataset of origin for the text. We decided to use the common online font \textit{Open Sans} and use a \textit{word-spacing} of 25\,px and a \textit{text-alignment} to the left. We use \textit{font-size} 24\,px and 22\,px and \textit{line-height} 3\,em and 1.9\,em, for XQuAD and MECO texts, respectively. These differences are due to MECO containing larger texts in comparison to those present in XQuAD. 
 
To generate the stimuli images we take screenshots of each of the texts for the resolution of $1280\times720$. These are then shown on the participant's screen. We pick this resolution for two reasons: (1) according to a survey of screen sizes around $77.2\%$ of users have the same or higher resolutions than $1280\times720$,\footnote{\url{https://gs.statcounter.com/screen-resolution-stats/desktop/worldwide}} and (2) lower resolutions are do not allow for the stimuli to be presented adequately. This way, we can use the resolution to generate the stimuli and we present the same resolution (and consequently the same spacing and font size) to all participants. The image is scaled if the participant uses a scaling factor in their browser or operating system. However, as long as the projected resolution is above the target, the image remains comparable. If the participant's screen resolution is lower than $1280\times720$, they will not be allowed to continue.

\subsection{Crowd-Sourcing Settings}\label{sec:crowd-sourcing}

We follow three strategies for collecting webcam eye-tracking data through crowd-sourcing settings in an online experiment: (1) reimbursed participation on \textit{Amazon Mechanical Turk}, (2) volunteer participation on \textit{Cognition}, and (3) controlled volunteer participation in the lab. In this section, we outline all three strategies. In total, we collected data from 600 participants (350 reimbursed participants ($\mu=34.5$ years, $\sigma=10.8$), 240 volunteer participants ($\mu=21.3$ years, $\sigma=5.5$), and 10 lab-controlled volunteers ($\mu=29.6$ years, $\sigma=4.2$).

\subsubsection{Reimbursed Participation}\label{sec:mturk-requirements}

For reimbursed participation, we set up the data collection on the crowd-sourcing platform of \textit{Amazon Mechanical Turk}\footnote{\url{https://www.mturk.com/}} (mTurk). We implement a stack that consists of using the cloud platform \textit{Heroku}\footnote{\url{https://www.heroku.com/}} to host our \textit{jsPsych} \citep{de_leeuw_jspsych_2015} experiment. We use \textit{psiTurk}\footnote{\url{https://github.com/NYUCCL/psiTurk}} \citep{gureckis_psiturk_2016} to handle the payment and posting of HITs\footnote{HIT (Human intelligent task) represents a single task performed by a participant on \textit{mTurk}. This term is defined by Amazon here: \url{https://www.mturk.com/worker/help}} on the \textit{mTurk} website. The code can be found online.\footnote{\url{https://github.com/tfnribeiro/WebQAmGaze}} 

The experiment sets are generated offline, consisting of a combination of existing plugins offered by \textit{jsPsych} extended with \textit{WebGazer} when needed. The sets are hosted online on a \textit{Heroku} server. We share this link with the workers from \textit{Amazon Mechanical Turk}, so they can complete the HIT through this link. For each set, containing 10 individual texts, we collect data from up to 9 unique participants. The set's data is then downloaded to a local machine for data processing. 

To participate in the HIT, the \textit{mTurk} participants need to comply with the following characteristics: (1) Participants are fluent in the language in which the texts are written;
(2) Participants are at least 18 years old; (3) Participants can read English (for the instructions); (4) Participants need to be on a laptop or desktop device with a webcam available; (5) Participants cannot have completed the same experiment before; (6) The screen needs to have a minimum reported resolution of $1280\times720$ to ensure the stimuli are presented consistently.

Additionally, we have further requirements in place to dissuade \textit{mTurk} participants from not completing the task correctly: (1) We only accept HITs that have an answer accuracy (i.e., the ratio of correct answers in the experiment) of $\geq50\%$; (2) Participants have a HIT approval rate of at least 95\% (this is only valid when workers have more than 100 HITs completed); (3) Participants have at least 5 HITs approved in their account and (4) Participants need to complete the task in less than 45 minutes. 

The participants received USD 7.00 for each HIT. Additionally, we offered bonuses based on the number of correct answers as a further incentive to pay attention during the task. These bonuses are offered based on the following criteria: USD 1.00 if they achieve at least 60\% answer accuracy, or USD 2.00 if they achieve at least 75\% answer accuracy.

We restrict the regions where the \textit{mTurk} experiment version is available to locations where the official language corresponds to the language of the texts being read in the experiment. The goal is to maximize the likelihood of getting fluent participants in countries where the language of the experiment is the primary language. The regions are set to US and Great Britain (for English); Spain, Mexico, Argentina, Colombia, Chile, Ecuador, Guatemala, Peru, and Venezuela (for Spanish); Germany and Austria (for German), and Turkey (for Turkish).\footnote{Since we accept any participants fluent in the given language -- not only native speakers -- we also tried recruiting participants by extending the list of countries, e.g., collecting German data in the US. However, we did not succeed in finding enough participants in this manner.}

The answers provided by the participants to each comprehension question in a set are corrected manually, except for the MECO texts, where the procedure is automatic since there is a binary choice. For XQuAD, answers are corrected manually based on how close to the original span they are. For some questions, we considered other responses than those marked in the original dataset due to the question being ambiguous or unclear. We ignore typos, as long as they are interpretable to the desired answer.

\subsubsection{Volunteer Participation}

As a second crowd-sourcing strategy we include a group of volunteer participants without reimbursement. The reason for including this second strategy is that for some of the languages, there were not enough workers available on \textit{mTurk} (i.e., German, Spanish, and Turkish). The volunteer-based version of the experiment is hosted on \textit{Cognition}\footnote{\url{https://www.cognition.run/}} and shared through the research networks of the authors. The main difference to the first setting is that no monetary reimbursement was offered to the volunteer participants. Therefore, there was also no time limit for completing the experiment. The experiment itself followed exactly the same structure and protocol as previously described.

\subsubsection{Lab-controlled Volunteer Participation}

Finally, we also collect data from a small group of participants at the University of Copenhagen. This is to ensure equal environment and light conditions for this set of 10 participants. We also use the same experiment implementation hosted on \textit{Cognition} for this set of participants. Through this controlled participation we can deduce the upper bound quality in eye-tracking accuracy achievable in our experiment setting. We use a MacBook Pro 13-inch (M1, 2020) with a camera sampling rate of 30 Hz.\\

\begin{figure*}[t]
    \centering
    \includegraphics[width=1\textwidth]{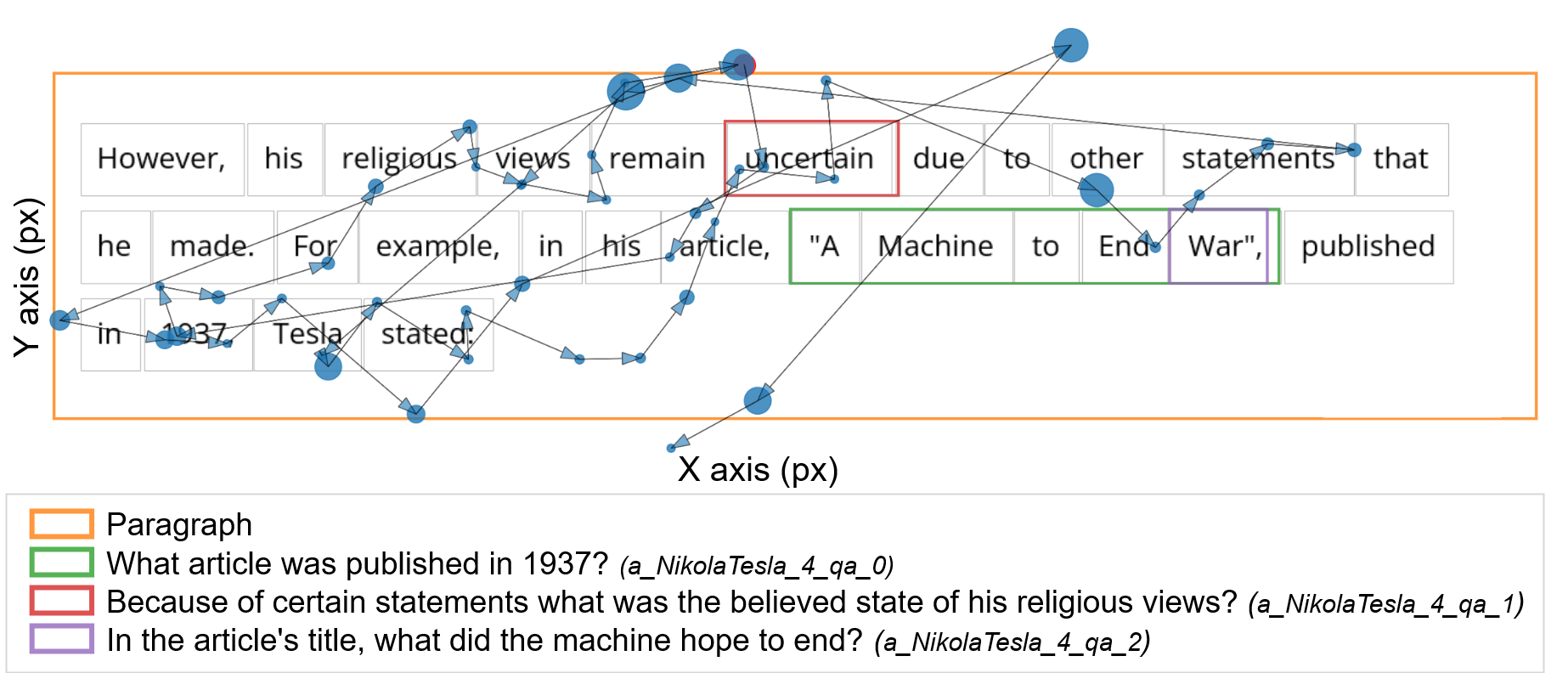}
    \caption{Example of the eye-tracking data and AOIs generated and data collected for XQuAD text \textit{NikolaTesla}, 5th paragraph. The blue dots show individual fixations, their size increases with longer fixation duration, and the arrows show the saccades and their direction between the fixations.
  The orange box is the AOI for the paragraph. In green, red, and purple are the target AOI passages to answer the three corresponding questions from the XQuAD dataset. Lastly, the grey boxes show the AOI for each word in the text. For this specific set, (\textit{mturk\_EN\_v10}), the question corresponds to the green box: \textit{``What article was published in 1937?''.}}
    %Participant A39SK1E6IMQBD5, EN_v10
    % This example is the first participant after filtering for EN_V10
    % The legends for the relevant passages follow a naming convention of a\_[NameOfParagraph]\_[ParagraphN]\_qa\_[QuestionN]. For this set in particular, \textit{mturk\_EN\_v10}, the question corresponds to \textbf{qa\_0}: \textit{What article was published in 1937?}, corresponding to the green highlight in the text.
    \label{fig:example_of_boundaries}
\end{figure*}

\begin{figure*}[t]
    \centering
    \includegraphics[width=1\textwidth]{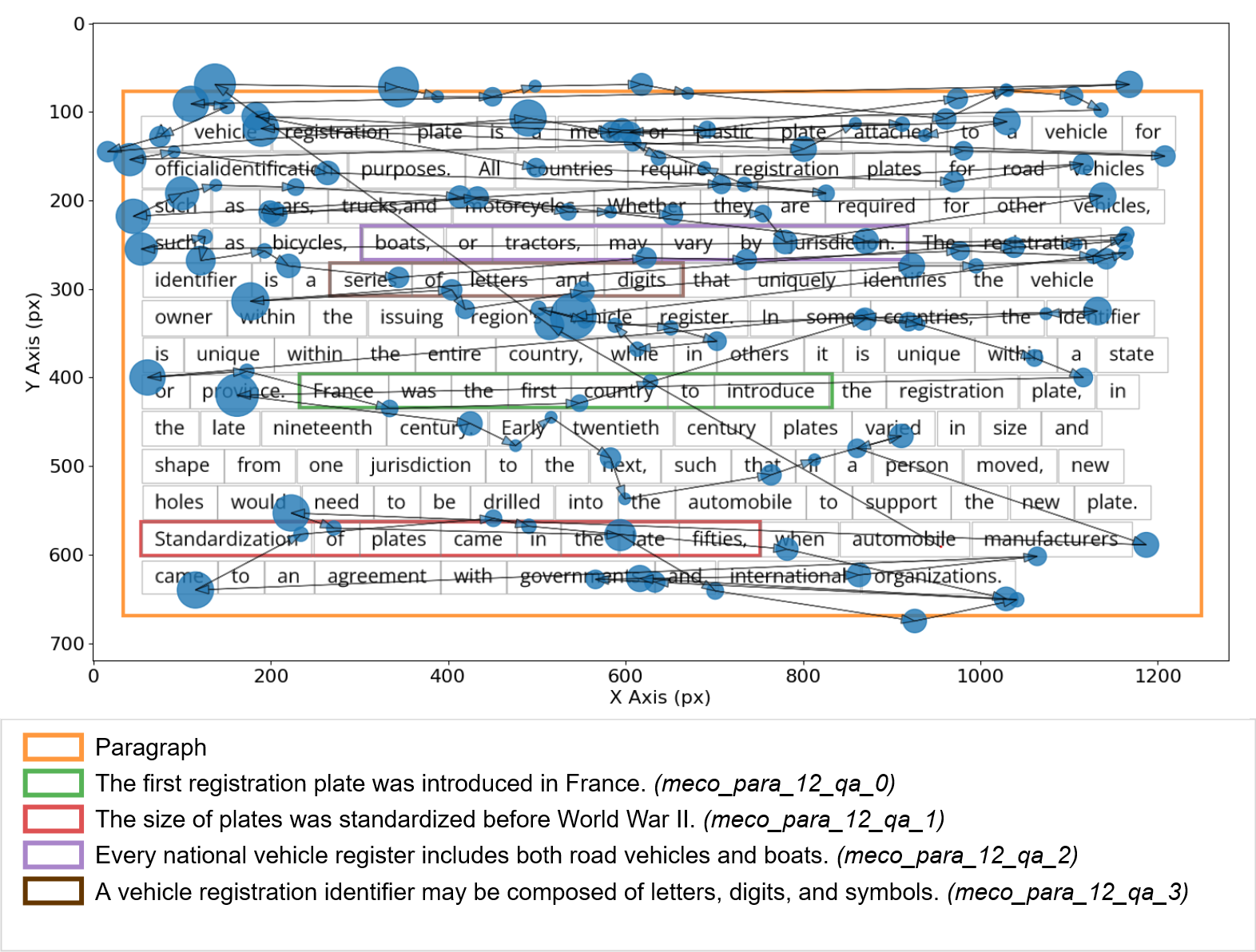}
    \caption{Example of the eye-tracking data and AOIs generated and data collected for MECO text \#12. The blue dots show individual fixations, their size increases with longer fixation duration, and the arrows show the saccades and their direction between the fixations.
  The orange box is the AOI for the paragraph. In green, red, purple and brown are the target AOI passages to answer the four corresponding True/False statements from the MECO dataset. Lastly, the grey boxes show the AOI for each word in the text. For this specific set, (\textit{mturk\_EN\_v10}), the statement corresponds to the red box: \textit{``The size of the plates was standardized before World War II.''.}}
    \label{fig:example_of_boundaries2}
\end{figure*}

\section{Data Processing}\label{preprocessing-steps}

In this section, we describe how the data was processed after collection. First, we filter the participant responses and the collected webcam eye movement data (Section \ref{sec:data-filtering}). Thereafter, we detail our approaches for fixation detection and merging (Section \ref{sec:fixation-detection}), as well as for identifying word-level areas of interest (Section \ref{sec:aoi-detection}).

\subsection{Data Filtering}\label{sec:data-filtering}

In the first step of data processing, the responses from the participants to the questions in all sets are corrected manually, with the exception of the MECO texts, where the procedure is automated since the answers are binary true/false choices. For XQuAD, answers are corrected manually based on how close to the original span they are. For some questions, we consider additional responses than the answer specified in the original dataset, due to ambiguities in the questions. We ignore typos, as long as they are interpretable to the original question.

\begin{figure*}[t]
        \centering
        \includegraphics[width=0.45\textwidth]{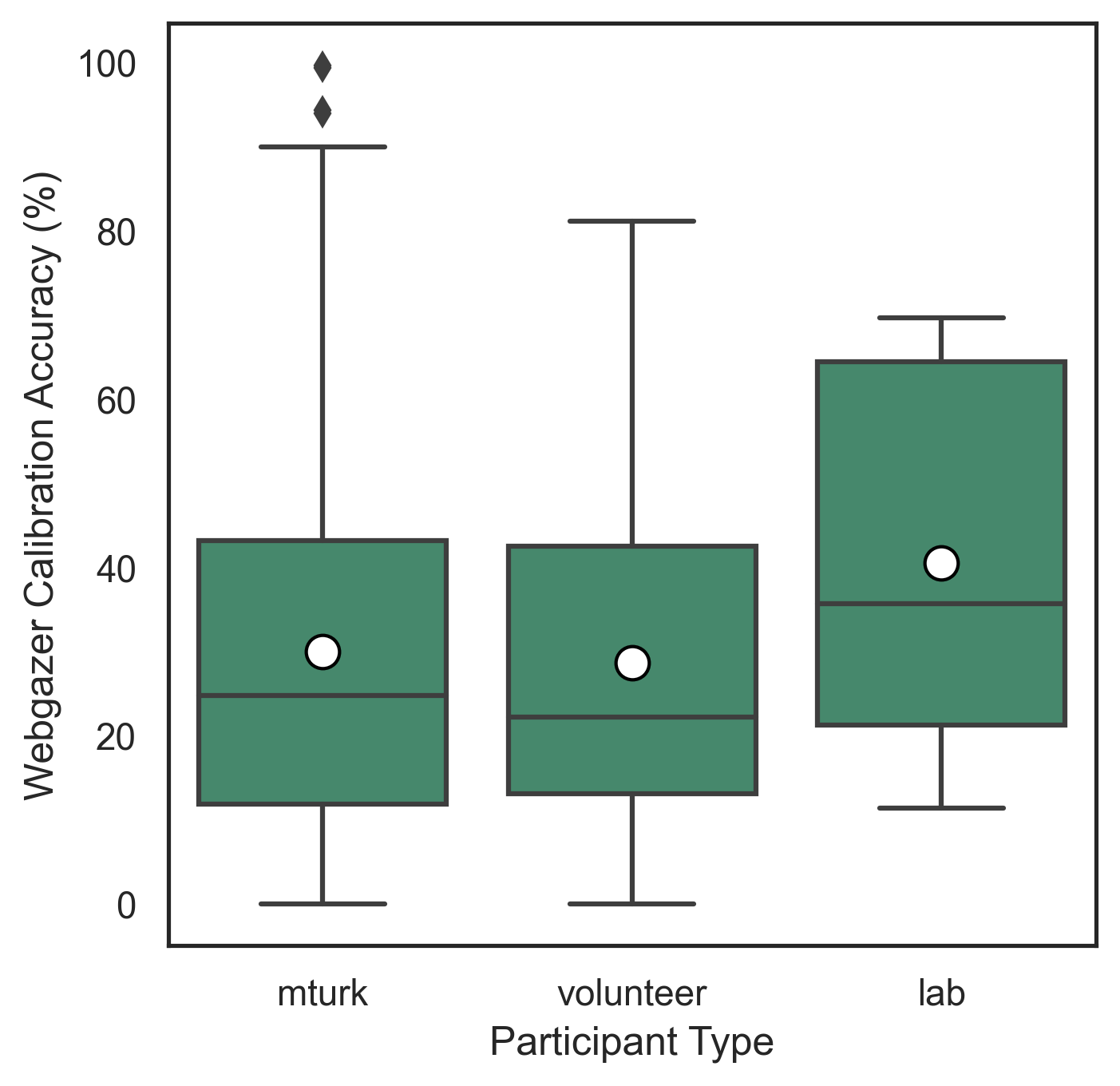}
        \caption{\textit{WebGazer} calibration accuracy (in \%) for all participant populations ($n=600$; 350 \textit{mTurk}, 240 volunteer, 10 lab). Medians are displayed as straight lines, means are shown as white dots.}
        \label{fig:compare-pop-calibration}
    \end{figure*}

Next, we compare the data quality from all three participant populations in terms of calibration accuracy. As we show in Figure \ref{fig:compare-pop-calibration}, the \textit{WebGazer} calibration accuracy is similar across \textit{mTurk} and volunteer participants, but higher for the lab-controlled volunteers. 
We first filter out participants who were not approved based on their number of correct questions (at least 50\% correct answers were required), which were $29.50\%$ of the total data collected. 
Additionally, we filter $4.26\%$ who experience an error with \textit{WebGazer}, resulting in either the targets or the gaze points not being stored correctly. We further remove $10.86\%$ based on a sample rate $<10\,Hz$, which is too low for linguistic processing, and $1.11\%$ due to their average calibration accuracy on the last validation step resulting in $0\%$, meaning that no point fell within the AOI across the 5-point validation. One participant is removed due to low screen resolution and lastly, 3 participants are removed due to taking more than 60 minutes to perform the experiment. After filtering with these criteria, we obtain data from $353$ participants out of the initial $600$. All following experiments and analyses in this work are performed on this filtered dataset.

\subsection{Fixation Detection and Gaze Point Filtering}\label{sec:fixation-detection}
We perform the following preprocessing steps on the gaze points obtained from the webcam recordings to transform them into fixation points.

In the first step, we implement a dispersion-threshold identification algorithm (I-DT) for fixation detection \citep{salvucci2000identifying}. 
First, we subtract the image coordinates $(i_x,i_y)$ (measured in pixels) from each gaze point $g=(g_x,g_y,g_t)$. For each of these initial gaze points, the $x,y$ coordinates refer to a location in the participant's screen. We perform the following operation to transform it into a resolution-independent gaze point $g'$:

\begin{equation}
    g' = (g_x-i_x,\, g_y-i_y,\, g_t)
\end{equation}

\noindent where $i_x,i_y$ are the coordinates of the top-left corner of image $i$ and $g_x,g_y$ is the estimation of the gaze location at time $g_t$, measured in milliseconds. The set of all resolution-independent gaze points is defined as $G'$, which refer to the image pixels of resolution $[0, 1280]$ on the $x$-axis and $[0, 720]$ on the $y$-axis.

Second, we merge the individual resolution-independent gaze points into fixations related to the reading process by defining a time window $w=150ms$ and a radius $r=32px$. For every point $g'_i=(x_i,y_i,t_i) \in G'$, we perform:

\begin{equation}
    f_i = (x,y,t_j)  = \frac{1}{N} \sum g'_j(x,y) 
\end{equation}

\begin{equation}
    \begin{split}
     \forall j>i\,:\, ||g'_j(x,y) - g'_i(x,y)|| &\\ < r \land g'_j(t)-g'_i(t) \le w &
    \end{split}
\end{equation}

\noindent where $f_i$ is the new fixation point created, and $N$ is the total number of points summed. Both $r$ and $w$ are empirically defined parameters. We then calculate the fixation duration to be the difference between the fixation time of two consecutive points, so that $d_i = t_i - t_{i-1}$ and set the first gaze point to have $d_0=0$. This point is kept and not filtered out when removing fixations based on duration. Additionally, we remove any data points that fall outside of the text target area, which is defined as the borders of the image containing the text with a tolerance threshold of $50px$ to allow for fixations related to reading that fall outside the text box due to lower calibration accuracy. 

As a final step, we also remove any fixations that are shorter than $50ms$. For webcams with sampling rates below 25Hz, this means that participants need to fixate the same region at least over the duration of two gaze points to be merged into a fixation. This process removes about 62.8\% of the total gaze samples after fixation detection. We use a value of $50ms$ as it provides a good balance between filtering non-fixation points and not losing too many data points. Lower values result in very little filtering due to the low frequency of \textit{WebGazer} and higher values result in over-filtering. It is important to note that both of these steps will significantly impact all the downstream tasks with eye movement measures derived from these fixations.  

We select this method for fixation detection as a simple initial approach that does not make strong assumptions about the quality of the 
gaze data collected and provides a sensible starting point for analysis.

% - word boundary detection
\subsection{Area of Interest Detection}\label{sec:aoi-detection}
\begin{table}[t]
    \centering
    \caption{Ratio of fixations falling within the different areas of interest (AOIs), reporting mean ($\mu$) and standard deviation ($\sigma$) calculated after fixation detection and gaze point filtering, including all fixations for all trials and all languages.}
        \begin{tabular}{lrr}
        \toprule
         & $ \mu $&   $\sigma$ \\
        \midrule
        Ratio of fixations within paragraph AOI  &  0.83 &  0.13 \\
        Ratio of fixations within Word AOIs       &  0.62 &  0.12 \\
        Ratio of fixations within Target AOIs     &  0.05 &  0.03 \\
        \bottomrule
        \end{tabular}
    \label{tab:ratio-fix-aoi}
\end{table}

No software is currently available to automatically generate word boundary boxes directly from stimulus images in conjunction with \textit{WebGazer} experiments. For this reason, we propose a method for automatic word boundary detection to retrieve rectangular boxes fitted around the individual word positions from the stimulus image file presented during the experiment. 
In the current work, we use three types of areas of interest (AOIs): (1) a \textit{paragraph AOI}, marking a box around the full text presented on the screen, (2) \textit{word-level AOIs}, i.e., an area of interest marked around each word, (3) \textit{target AOIs}, marking a box around each target span relevant for answering the question. We present the ratios of the number of fixations that fall within the different AOIs in Table \ref{tab:ratio-fix-aoi}.

We use \textit{pytesseract}\footnote{\url{https://pypi.org/project/pytesseract/}}, a Python wrapper for the OCR engine \textit{libtesseract}, to retrieve the AOIs given the images used as stimuli for the experiment. Based on the bounding boxes detected we create an area of interest for each of the words, by taking the originally detected bounding boxes and expanding them to ensure that they provide some margin of error, but without overlapping with other words. These values were tuned empirically based on the values set in the CSS when creating the images.

Two examples of the calculated AOIs and eye-tracking data from two random participants are shown in Figures \ref{fig:example_of_boundaries} and \ref{fig:example_of_boundaries2}. As can be seen, the eye-tracking data from the participant plotted in Figure \ref{fig:example_of_boundaries2} is more accurate than the one in Figure \ref{fig:example_of_boundaries}.
Once again, it is worth noting that the word-level eye movement measures lack in accuracy. To illustrate, the height of all words is 40px, and the width is 48px for the word ``in" and 150px for the word ``competitive". Therefore, an error of 100px is indeed an error that can go across words (horizontally) or lines (vertically).

\section{Data Analysis}

We analyze the collected \textsc{WebQAmGaze} data in various steps. First, we present basic dataset statistics (Section \ref{sec:data-stats}). Next, we compare the webcam data to high-quality eye-tracking recordings (Section \ref{sec:meco-comparison}). Finally, we explore with preliminary methods how the webcam data can be used for explainable AI (Section \ref{sec:explainable-ai}).

\subsection{Dataset Statistics}\label{sec:data-stats}

\begin{figure*}[t]
     \centering
     \begin{subfigure}[b]{0.43\textwidth}
         \centering
         \includegraphics[width=\textwidth]{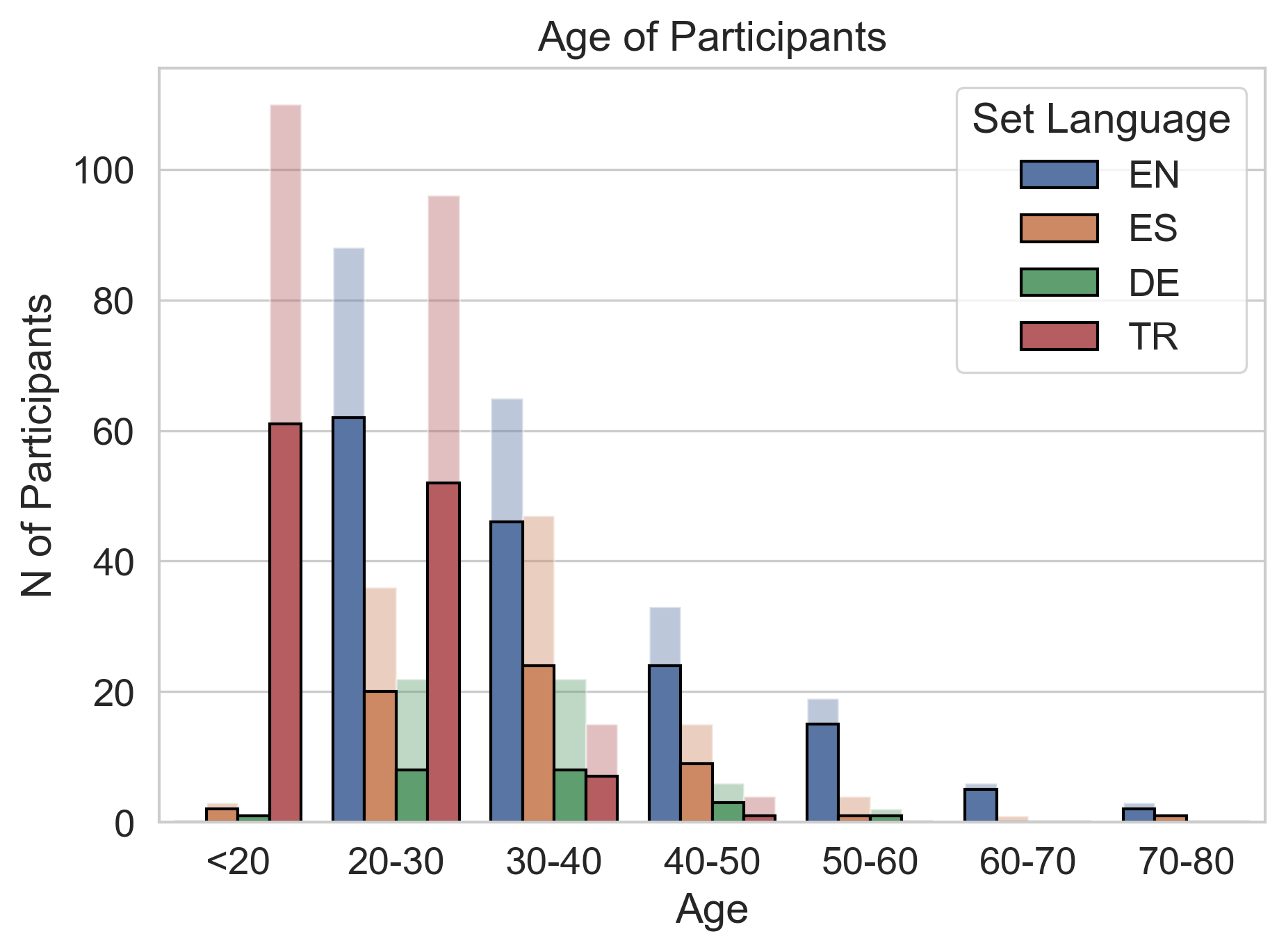}
         \caption{Participants' age distribution.}
         \label{fig:participants-age}
     \end{subfigure}
     \hfill
     \begin{subfigure}[b]{0.43\textwidth}
         \centering
         \includegraphics[width=\textwidth]{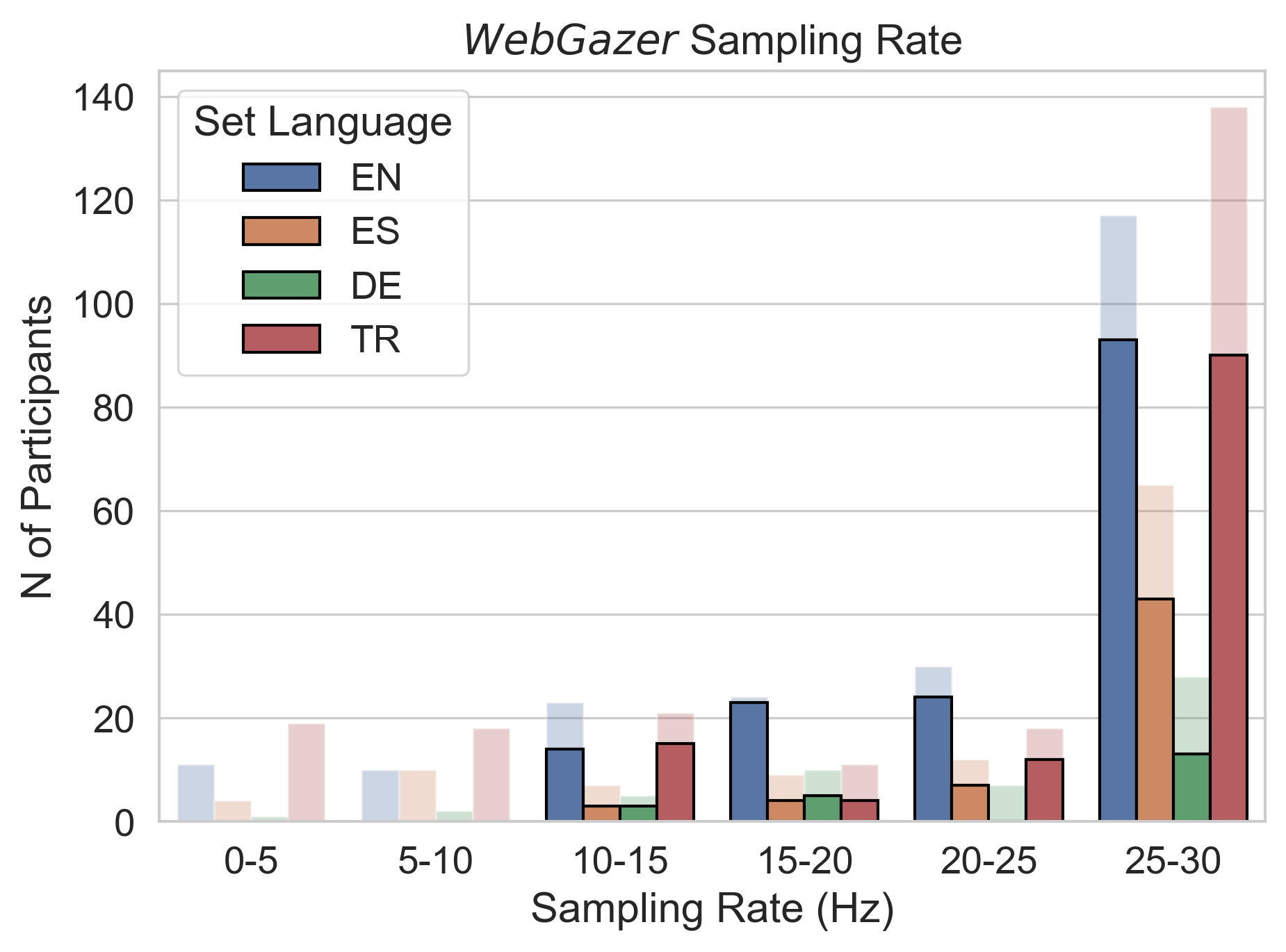}
         \caption{Sampling rate distribution.}
         \label{fig:sample-rate}
     \end{subfigure}
     \hfill
        \caption{Participants' age and \textit{WebGazer} sampling frequency distribution. Bars in lighter colors show the full data before filtering ($n=600$), darker bars show the filtered data ($n=353$).}
        \label{fig:experiment-results-bar}
    \end{figure*}

\begin{figure*}[t]
     \centering
     \begin{subfigure}[b]{0.43\textwidth}
         \centering
         \includegraphics[width=\textwidth]{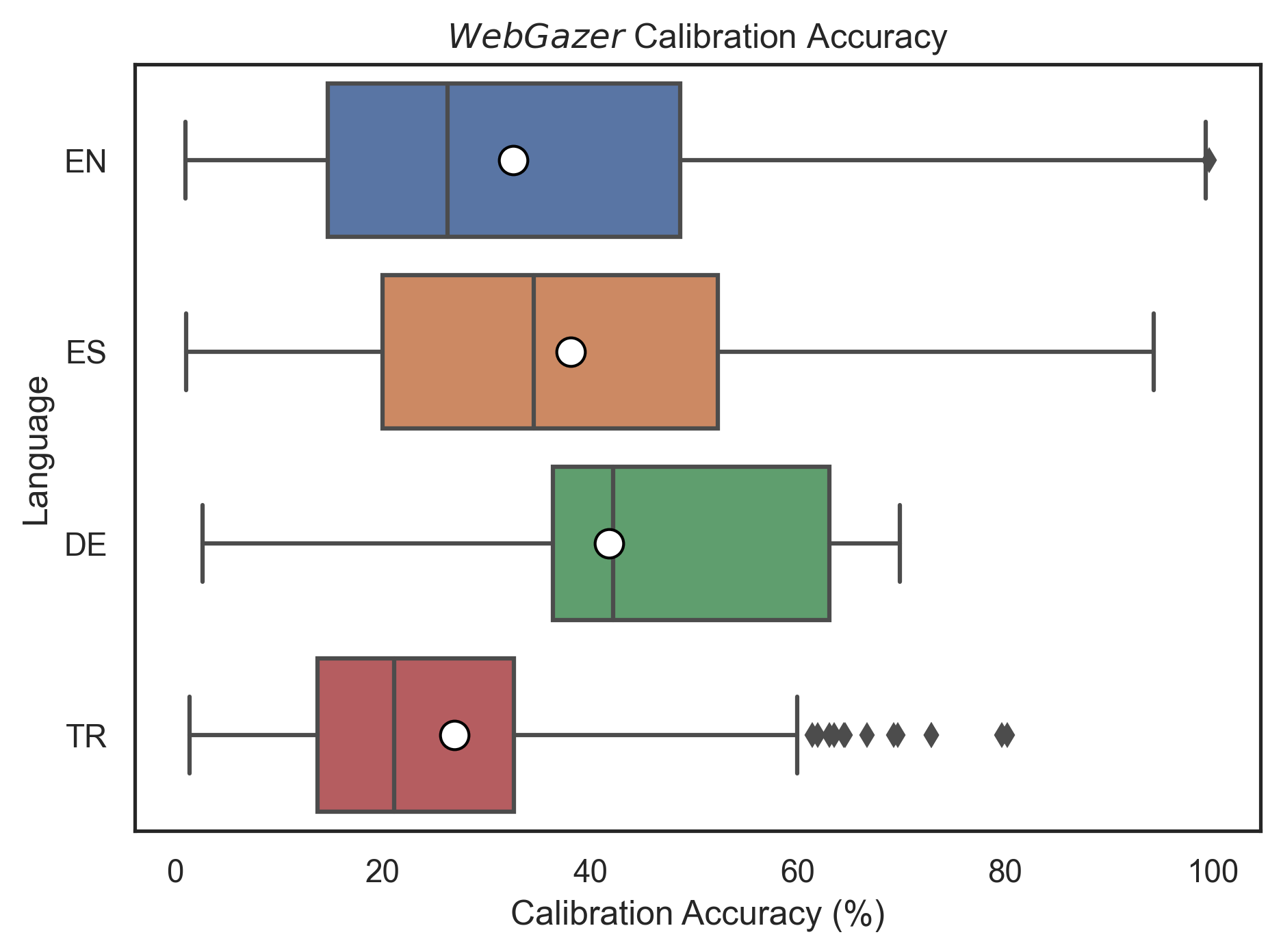}
         \caption{\textit{WebGazer}'s calibration accuracy for each language.}
        %of 5 points (100 ROI radius)performed at the start of the experiment}
         \label{fig:roi-acc}
     \end{subfigure}
     \hfill
     \begin{subfigure}[b]{0.43\textwidth}
         \centering
         \includegraphics[width=\textwidth]{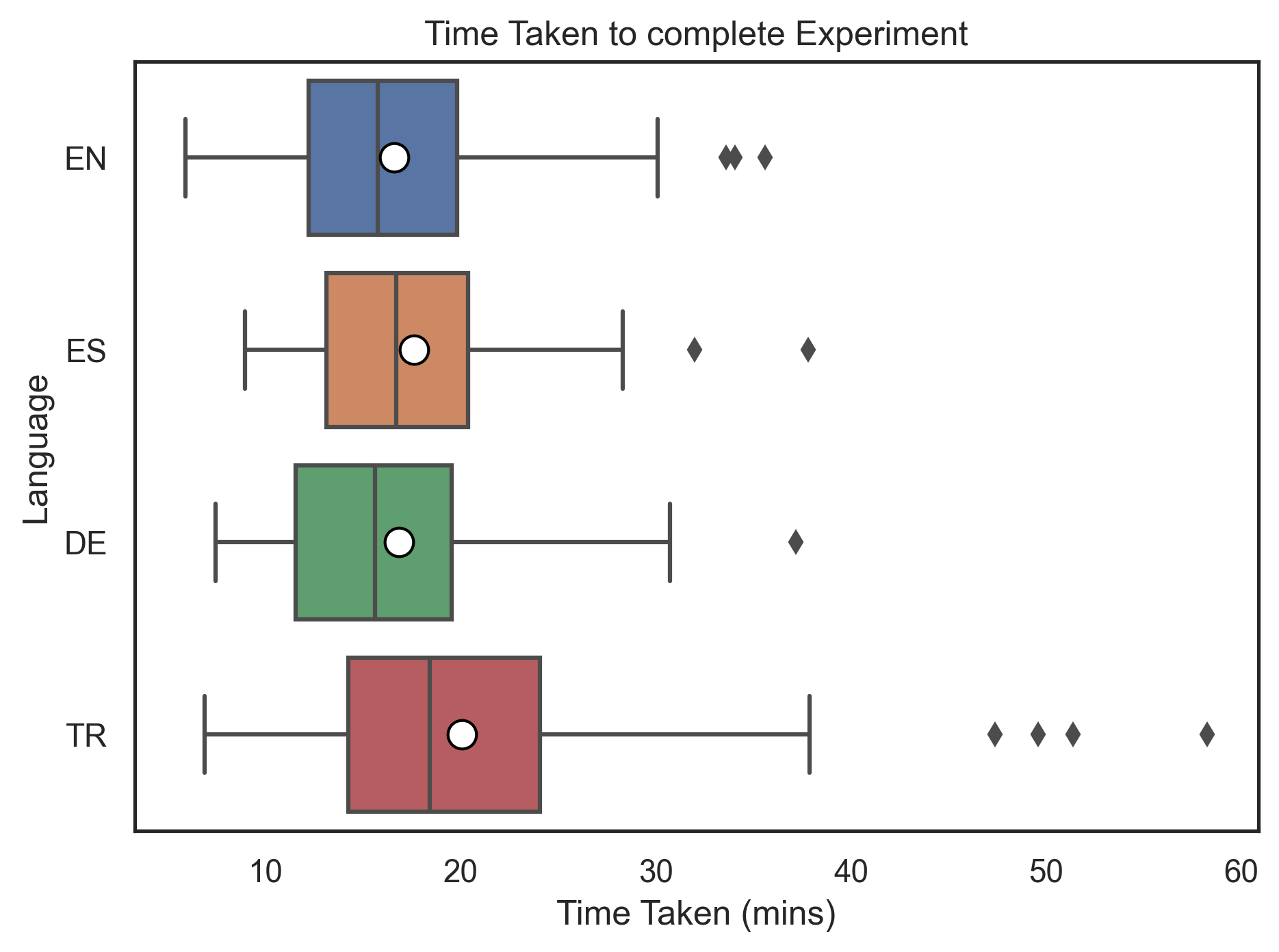}
         \caption{Total time spent completing the experiment, separated by language.}
         \label{fig:time_spent_experiment}
     \end{subfigure}
     \hfill
        \caption{\textit{WebGazer} calibration accuracy and total experiment duration per language.}
        \label{fig:experiment-results-box}
    \end{figure*}

We collect data from a total of 600 participants. After performing the filtering steps described in Section \ref{preprocessing-steps}, we retain data from 353 participants, 154 for English, 57 for Spanish data, 21 for German data, and 121 for Turkish. We report the participants' age distribution  ($\mu=29.47$ years, $\sigma=11.74$) and the sampling rate of the webcams measured ($\mu=25.39$ Hz , $\sigma=5.67$) in Figure \ref{fig:experiment-results-bar}. The \textit{WebGazer}'s calibration accuracy after the last 5-point validation step ($\mu=32.10$ \%, $\sigma=22.28$) and the time taken to complete the full experiment ($\mu=17.97$ mins, $\sigma=7.37$) are shown in Figure \ref{fig:experiment-results-box}.

To complement the calibration accuracy as a metric of data quality, we also report the average offset from the calibration data calculated as the average distance from each validation point (in pixels). Table \ref{tab:median_offset_per_resolution} shows the average offset for each of the screen resolutions in our participant groups. 
Figures \ref{fig:good-validation} and \ref{fig:bad-validation} show examples of good and bad validation results.

\begin{table}[ht]
    \centering
    \begin{tabular}{lrrrrrr}
    \toprule
     \textbf{Resolution} & \textbf{Participants} &    \textbf{Min} &     \textbf{Max} &    \textbf{Mean} &    \textbf{Std} &  \textbf{Median} \\
    \midrule
     1366x768 &                 95 &  30.97 &  320.95 &  112.96 &  52.95 &  100.15 \\
     1536x864 &                 86 &  23.06 &  593.23 &  109.24 &  75.18 &   87.78 \\
    1920x1080 &                 40 &  29.59 &  207.75 &  101.45 &  45.14 &   75.71 \\
     1440x900 &                 37 &   0.03 &  181.36 &   81.42 &  46.02 &   66.38 \\
     1280x720 &                 24 &   0.01 &  276.39 &   82.82 &  57.22 &   65.22 \\
     1600x900 &                 18 &  26.57 &  295.94 &  137.84 &  61.48 &  116.82 \\
     1280x800 &                  9 &   0.29 &  151.92 &   76.57 &  43.14 &   67.27 \\
    2560x1440 &                  6 &  55.50 &  217.86 &  105.28 &  61.73 &   73.77 \\
    \bottomrule
    \end{tabular}
    \caption{Average offset error in pixels for the different screen resolutions and number of participants in the \textsc{WebQAmGaze} dataset.}
    \label{tab:median_offset_per_resolution}
\end{table}

\subsection{Comparison to High-Quality Eye-Tracking Data}\label{sec:meco-comparison}

\begin{figure*}[t]
    \centering
    \includegraphics[width=.95\textwidth]{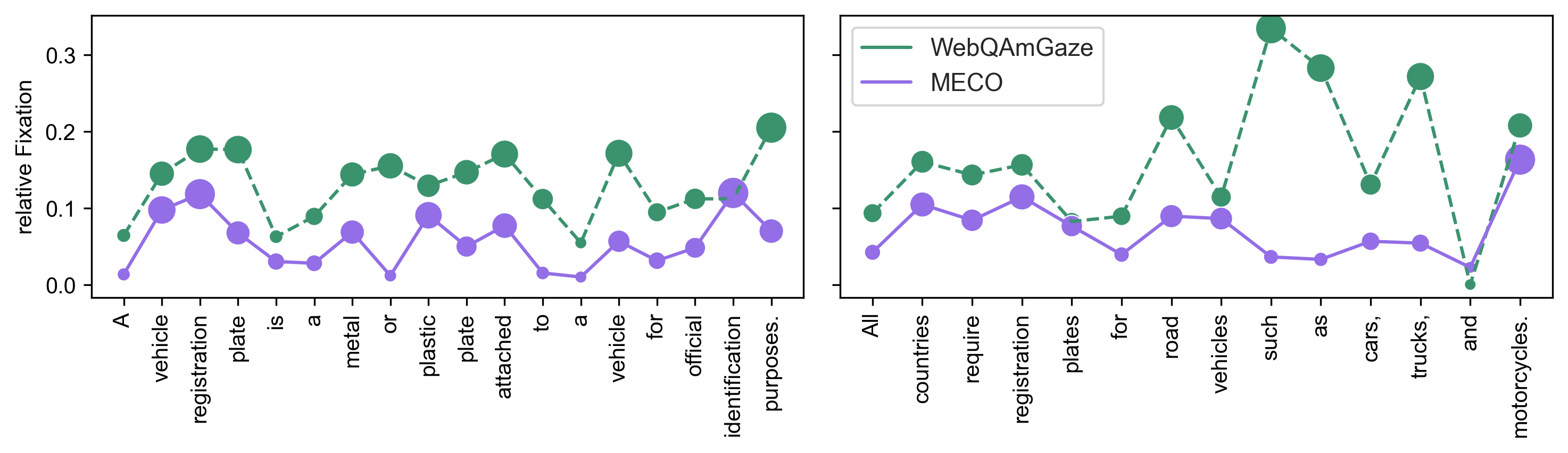}
    \caption{Comparison between relative fixation duration patterns from MECO and \textsc{WebQAmGaze} on two individual sentences, average across participants.}
    \label{fig:sen_comparison}
\end{figure*}

\begin{figure}[t]
    \centering
    \begin{subfigure}[b]{0.43\textwidth}
         \centering
         \includegraphics[width=\textwidth]{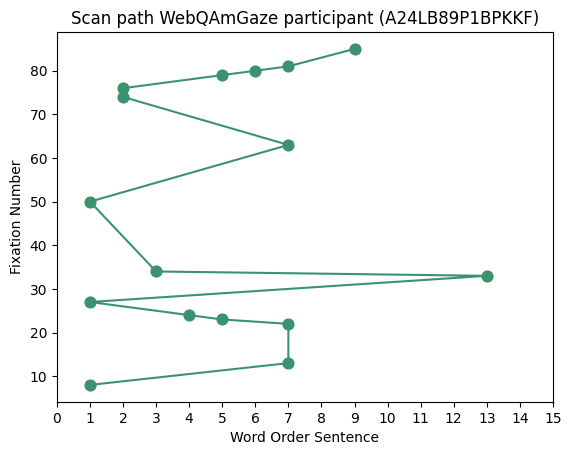}
         \label{fig:webqam-scan}
     \end{subfigure}
     \hfill
     \begin{subfigure}[b]{0.43\textwidth}
         \centering
         \includegraphics[width=\textwidth]{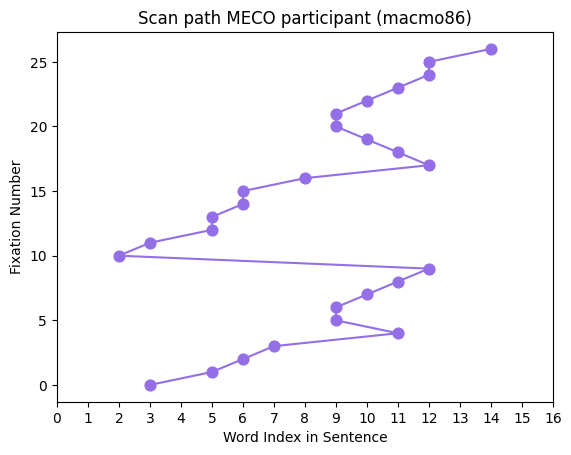}
         \label{fig:meco-scan}
     \end{subfigure}
    \caption{Scanpath examples for the sentence ``It is involved in data gathering and analysis, research, field projects, advocacy, and education.'' (MECO, Text 11). \textsc{WebQAmGaze} on the left and MECO on the right. The x-axis corresponds to the word index in the sentence, and the y-axis to the fixation index in the sentence.}
    \label{fig:scanpath}
\end{figure}

To assess the quality of the webcam recordings, we compare the webcam recordings for the included MECO texts with the original MECO eye-tracking data by \citet{siegelman_expanding_2022_meco}. The MECO dataset was recorded in 13 different laboratories (one per language) around the world using EyeLink trackers with a sampling rate of 1000\,Hz. Participants used a chin rest and a head restraint to minimize head movements. Thus the data is expected to be of much higher quality than \textsc{WebQAmGaze} with respect to comparability between participants and eye-tracker accuracy.

We compute the total reading time (TRT) and the number of fixations (nfix), both averaged across all participants and words. TRT is calculated as the summed duration of all fixations on a word. Note that we only include words that were fixated at least once in the calculation of TRT, whereas for nfix we include all words, starting with a count of $0$. 

For a comparison independent of the absolute fixation duration measurements, we also compute the Spearman correlation coefficients between the relative fixation durations of both datasets. To compute relative fixation duration (RFD), the total fixation time (TRT) per word is divided by the sum of all TRTs in the respective sentence for all individual participants for each text, as proposed by \citet{hollenstein-beinborn-2021-relative}. We omit words that have not been fixated in the entire dataset.

% --- NEW ---
\begin{table}[t]
\centering
\caption{Comparison between MECO texts in \textsc{WebQAmGaze} and original MECO data. Text ID, number of participants (\# parts.), mean total reading time (TRT, measured in \textit{ms}), and number of fixations (nfix) averaged across all participants and words. Note that we only include fixated words in the calculation for TRT. The last columns show the Spearman correlation coefficient $\rho$ between relative fixation duration averaged across participants in both datasets. All results are significant ($p<0.01$).}
\label{tab:comparison}
\begin{tabular}{l|rr|cr|crr|crr}
\toprule
 &  &  & \multicolumn{2}{c|}{\textbf{MECO}} & \multicolumn{3}{c|}{\textbf{WebQAmGaze}} & \multicolumn{3}{c}{\textbf{WebQAmGaze} (20\%)} \\
Lang & ID & \# parts. & TRT {[}ms{]} & nfix & TRT {[}ms{]} & nfix & $\rho$ & TRT {[}ms{]} & nfix & $\rho$ \\\midrule
\multirow{3}{*}{{EN}} & 3  & 40 & 244 & 1.18 & 365 & 2.10 & \textcolor{black}{0.58} & \textcolor{black}{305} & \textcolor{black}{1.50} & \textcolor{black}{0.51}\\
                      & 7  & 13 & 226 & 1.09 & \textcolor{black}{284} & \textcolor{black}{1.09} & \textcolor{black}{0.64} & \textcolor{black}{291} & \textcolor{black}{1.19} & \textcolor{black}{0.60}\\
                      & 11 & 75 & 260 & 1.22 & \textcolor{black}{370} & \textcolor{black}{1.64} & \textcolor{black}{0.59} & \textcolor{black}{384} & \textcolor{black}{1.87} & \textcolor{black}{0.60}\\ 
                      & 12 & 26 & 210 & 1.02 & \textcolor{black}{307} & \textcolor{black}{1.61} & \textcolor{black}{0.63} & \textcolor{black}{293} & \textcolor{black}{1.59} & \textcolor{black}{0.60} \\\midrule
                   ES & 12 & 57 & 245 & 1.22 & \textcolor{black}{333} & \textcolor{black}{1.43} & \textcolor{black}{0.68} & \textcolor{black}{329} & \textcolor{black}{1.6} & \textcolor{black}{0.71}\\\midrule
\multirow{2}{*}{{DE}} & 1 & 6 & 313 & 1.49 & \textcolor{black}{530} & \textcolor{black}{1.84} & \textcolor{black}{0.50}  & \textcolor{black}{755} & \textcolor{black}{2.60} & \textcolor{black}{0.43} \\
                      & 12 & 15 & 301 & 1.46 & \textcolor{black}{284} & \textcolor{black}{1.43} & \textcolor{black}{0.66} & \textcolor{black}{290} & \textcolor{black}{1.47} & \textcolor{black}{0.65}\\ 
\midrule
\multirow{2}{*}{{TR}} & 7 & 117 & 363 & 1.8 & \textcolor{black}{508} & \textcolor{black}{2.95} & \textcolor{black}{0.76} & \textcolor{black}{489} & \textcolor{black}{2.77} & \textcolor{black}{0.72}\\ 
                      & 11 & 4 & 403 & 1.95 & \textcolor{black}{286} & \textcolor{black}{1.86} & \textcolor{black}{0.35} & \textcolor{black}{295} & \textcolor{black}{2.04} & \textcolor{black}{0.38}\\\bottomrule
\end{tabular}
\end{table}

The results are presented in Table \ref{tab:comparison}. In addition to a comparison including all trials, we add the results of filtering out the \textsc{WebQAmGaze} data with $<20\%$ \textcolor{black}{calibration} accuracy.

Compared to MECO, in the \textsc{WebQAmGaze} data we see an increase in TRT and a slight increase in the number of fixations, particularly for English and Spanish. We assume this is caused by the overall lower \textcolor{black}{calibration} accuracy and drastically lower sampling rates in the webcam data. 
%When assigning fixations to individual words as some fixations are aggregated to one word instead of being split across neighboring words. 
We see significant correlation values between the \textcolor{black}{RFD} of both datasets. %mostly around $0.55$, which is substantially higher than the correlation between MECO and first-layer attention as shown in \citet{brandl-hollenstein-2022-every}. 
The correlations for text $1$ in German and text $11$ in Turkish are substantially lower than the other values. Note that here only 6 and 4 participants have been recorded for these sets, respectively. \textcolor{black}{On the other hand, for text 7 in Turkish the correlation between RFD in both datasets is very strong.} This text was read by 117 participants.
Furthermore, when applying the $<20\%$ \textcolor{black}{calibration} accuracy filter, some of the TRT values increase even more while the correlation remains mostly similar. 
These results are an optimistic sign about the data quality of webcam eye-tracking, suggesting that higher webcam \textcolor{black}{calibration} accuracy is not a necessary requirement for a stronger correlation to high-quality eye-tracking devices. \textcolor{black}{However, a comparison of the webcam eye-tracking data to a wider range of high-quality eye-tracking datasets is recommended to validate the correlation results presented in Table \ref{tab:comparison} in the future. Dissimilarities between datasets can arise not only from differing eye-tracker accuracies and sampling rates but also from varying stimulus presentations and reader populations.}

In Figure \ref{fig:sen_comparison}, we show patterns of relative fixation \textcolor{black}{duration} of two-handpicked English sentences for \textsc{WebQAmGaze} and MECO. Both follow a similar pattern, where \textsc{WebQAmGaze} shows substantially higher deviations than MECO. On the sentence to the right, we can see that the fixation-to-word mapping in the webcam data is imperfect, as it shows a high RFD for \emph{such} which might actually correspond to the prior word \emph{vehicle}. \textcolor{black}{We further plot an example of two participants' scan path from MECO and \textsc{WebQAmGaze} dataset to illustrate spatial effects (see Figure \ref{fig:scanpath}). This shows that there is some fixation information missing in the webcam data, which can be assigned to a line below or above, but the general path followed remains similar in both cases.}

\begin{figure}[t]
    \centering
    \includegraphics[width=0.63\textwidth]{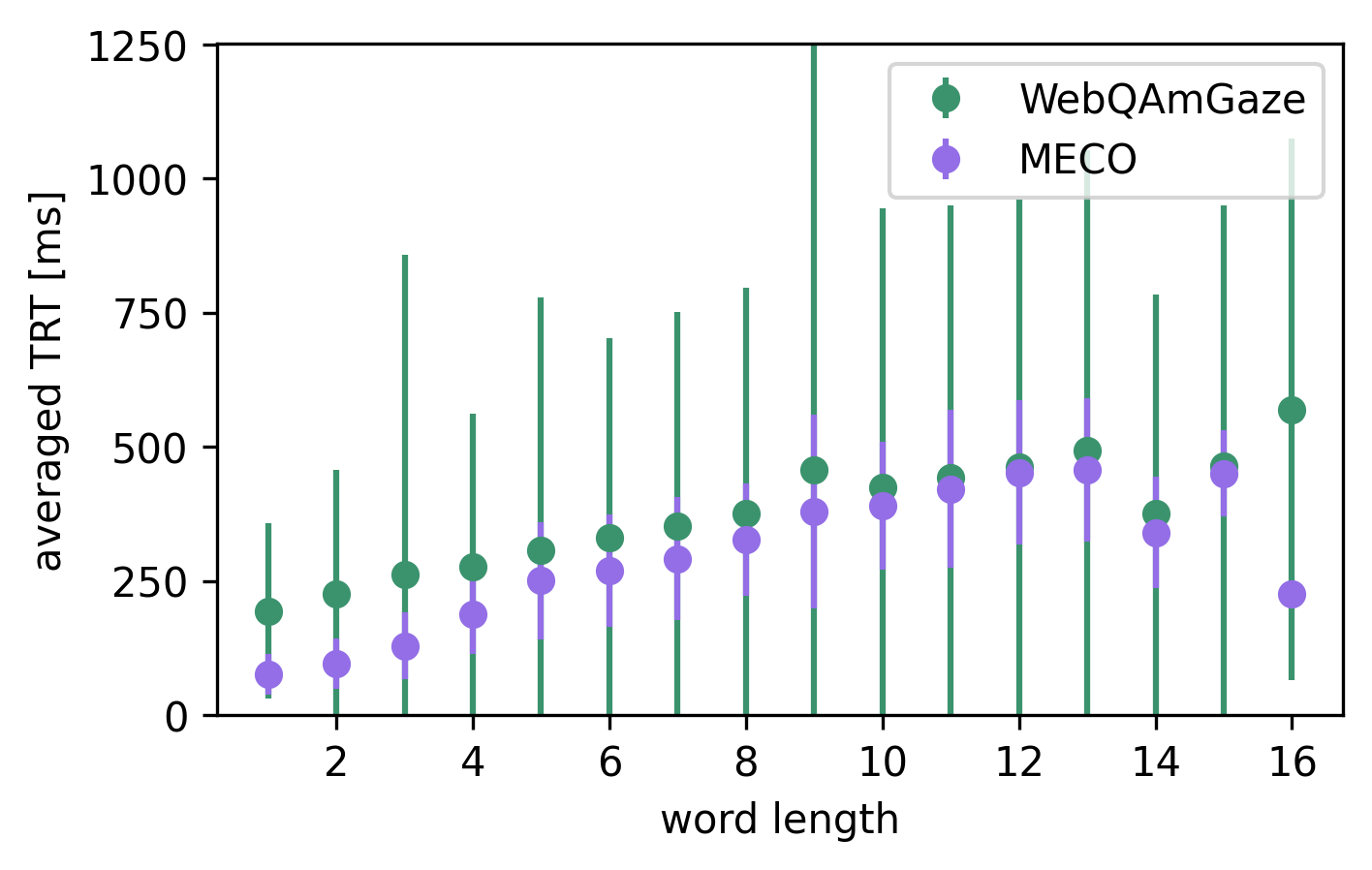}
    \caption{Averaged total reading time (TRT) in milliseconds per word length (i.e., the number of characters) for \textsc{WebQAmGaze} (all English sentences in the NR condition) and MECO (all English texts). TRT is averaged across tokens and participants, standard deviation is shown after averaging across participants.}
    \label{fig:wordlen}
\end{figure}

Moreover, in Figure \ref{fig:wordlen}, we show TRT averaged across participants and tokens based on their \textcolor{black}{word length for all English words in the NR paradigm of \textsc{WebQAmGaze}, against the English MECO data}. \textcolor{black}{We see an increase in TRT for longer words, which is in line with the well-studied word length effect in reading \citep{just1980theory} and also aligns with results from other high-quality eye-tracking-while-reading datasets (e.g., \citet{cop2017presenting,hollenstein2018zuco}).}

\subsection{Towards Eye Movement Rationales for Explainable AI} \label{sec:explainable-ai}

Current state-of-the-art machine learning models for language processing are still mostly black boxes composed of deep neural networks, making it difficult to comprehend why certain decisions are made in their outputs. Explainable AI models aim to provide justifications for the decisions made by a model, often based on human annotations or rationales \citep{sogaard2021explainable}. A rationale can be defined as a justification for a particular decision taken by a model for a specific NLP task. For example, in a question-answering model, a rationale is the target span in a given text relevant for correctly answering a question. 

Datasets such as ERASER \citep{deyoung-etal-2020-eraser} provide 
a methodology that can be used to compare and evaluate how explainable different model rationales are. They introduce metrics that aim to capture how well rationales provided by computational models align with human rationales (see \citet{wiegreffe2021teach} for a review).
However, collecting these rationale annotations is time-consuming and involves conscious task-solving. Therefore, they 
include human subjectivity and biases \citep{chiang2022reexamining}.
% show that depending on the qualification of the annotators and what instructions are given to them, different annotations are produced, despite the same underlying task. They criticize the somewhat lack of details sometimes provided by datasets containing rationales and call for careful interpretation of the results when compared to these datasets.

We hypothesize that eye movement information can be used to extract rationales without the need for annotators to manually mark which spans are relevant. The conscious, time-consuming process can be replaced by gaze information collected while reading, a more efficient process grounded in human attention. As an initial validation for this hypothesis, we investigate whether fixation-based eye movement measures extracted during reading are indicative of correctly answering the corresponding question about the text. In Section \ref{sec:significance-testing}, we define the eye movement measures and perform a set of statistical significance tests, followed by a machine learning approach in Section \ref{sec:classifier} classifying whether the correct answer is given or not.

\subsubsection{Significance Testing}\label{sec:significance-testing}
We evaluate if there is a significant difference in answering the question of a given trial correctly or incorrectly \textcolor{black}{for 7 eye movement measures. These measures are (1) number of fixations on target AOI, (2) number of fixations (total fixation points during the trial), (3) target fixation ratio (the ratio of (1)/(2)), (4) TRT on paragraph AOI, (5) TRT on target AOI, (6) target TRT ratio (the ratio of (5)/(4)), and (7) total trial time (the full time spent on a trial screen).}
%To test whether the hypothesis that eye movement measures are predictive of answer correctness can be confirmed, 
We perform independent t-tests for both reading tasks (NR and IS), by grouping participants by whether they responded correctly or incorrectly to the question of a given trial. 
For each of the t-tests, we group participants by the task performed (NR, IS), and test if these seven eye movement measures are significantly different between trials that get a correct answer versus trials that receive an incorrect answer to the question asked after reading the text.

\begin{table}[t]
\centering
\caption{Significance testing for \textbf{all languages together}. $p$-values obtained for independent t-tests performed by separating the groups into correct or incorrect responses given in a trial. Results are aggregated over the NR or IS task. The total number of trials is %$n=3530$ 
$n=1765$ and $n=1060$ for each task without and with the 20\% \textcolor{black}{calibration} accuracy filter, respectively. 
%, for feature (6) the number is $n=3370$, due to some lack of TRT text for some participants. 
%$n=2120$ ($1025$ for each task) with the 20\% filter. 
%feature (6) $n=2050$ for the same reason. 
\textcolor{black}{Significant results ($p<0.05$) are marked with *. Results marked with ** remain significant under the Bonferroni correction ($p<0.007$).}}
\label{tab:t-tests}
\begin{tabular}{l|rr|rr}
\toprule
\textbf{Feature} & \multicolumn{2}{c|}{\textbf{NR}} & \multicolumn{2}{c}{\textbf{IS}} \\
 & all & 20\% filter & all & 20\% filter \\\midrule
(1) Fixations on target AOI        & 0.775 & 0.634 & * 0.027 & * 0.039 \\
(2) Number of fixations             & 0.191 & 0.311 & ** $<0.001$  & ** $<0.001$  \\
(3) Target fixation ratio & 0.069 & 0.185 & 0.677 & 0.461 \\
(4) TRT on paragraph AOI   & 0.384 & 0.445 & ** $<0.001$ & ** $<0.001$  \\
(5) TRT on target AOI      & 0.636 & 0.726 & * 0.045 & * 0.028 \\
(6) Target TRT ratio      & 0.053 & 0.165 & 0.931 & 0.588 \\
(7) Total trial time    & * 0.028 & 0.067 & ** $<0.001$  & ** $<0.001$ \\\bottomrule
\end{tabular}
\end{table}

The results are shown in Table \ref{tab:t-tests} for all languages. We present results for two settings, all trials and trials filtered from participants with $>20\%$ \textcolor{black}{calibration} accuracy. For all trials, we find a significant difference for NR only in total trial time (7). The results are stronger for IS, where all measures except (3) and (6) are significant. Given the nature of the tasks -- the NR task does not guide participants in any way -- these results match our expectations. 

\textcolor{black}{Additionally, only total trial time (7) has a significant effect on both tasks, when no filter is applied. Looking at the total trial time means of the two groups in NR, the group that has correct answers takes longer to respond (incorrect answer: $\mu_1=37.6 s$ vs correct answer: $\mu_2=40.9 s$, while in IS the reverse is true (Wrong Group: $\mu_1=28.9 s$ vs Correct Group: $\mu_2=23.2 s$)}, shorter total trial time times affects the correctness of the answers. This lines up with our hypothesis, as in the NR task, participants need to read attentively as they are not aware which question will be asked, while in the IS task participants are incentivized to find only the relevant information.
%(incorrect answer: $\mu_1=37558ms$ vs correct answer: $\mu_2=40925ms$, while in IS the reverse is true (Wrong Group: $\mu_1=28899 ms$ vs Correct Group: $\mu_2=23226ms$)

Analogous to the results in Section \ref{sec:meco-comparison}, applying the 20\% \textcolor{black}{calibration accuracy} filter does not make the differences more apparent. On the contrary, for NR no feature yields a significant difference, while for IS the results remain unchanged. Note, however, that the sample size after filtering is reduced. Again this hints towards the fact that merely filtering by calibration accuracy does not automatically improve data quality and that noise in the data is also hidden in other data quality aspects such as the sampling rate of the camera. \textcolor{black}{Nevertheless, the results presented in Table \ref{tab:t-tests} show that for information-seeking reading, fixation on target AOI, number of fixations, TRT on paragraph AOI and target AOI, and total trial time are significant predictors of whether a reader will answer the question correctly or not.}

In Appendix \ref{app:ttests}, we present additional results for each language individually, where the results show similar patterns.

\begin{figure}[!t]
    \centering
    \includegraphics[width=0.58\textwidth]{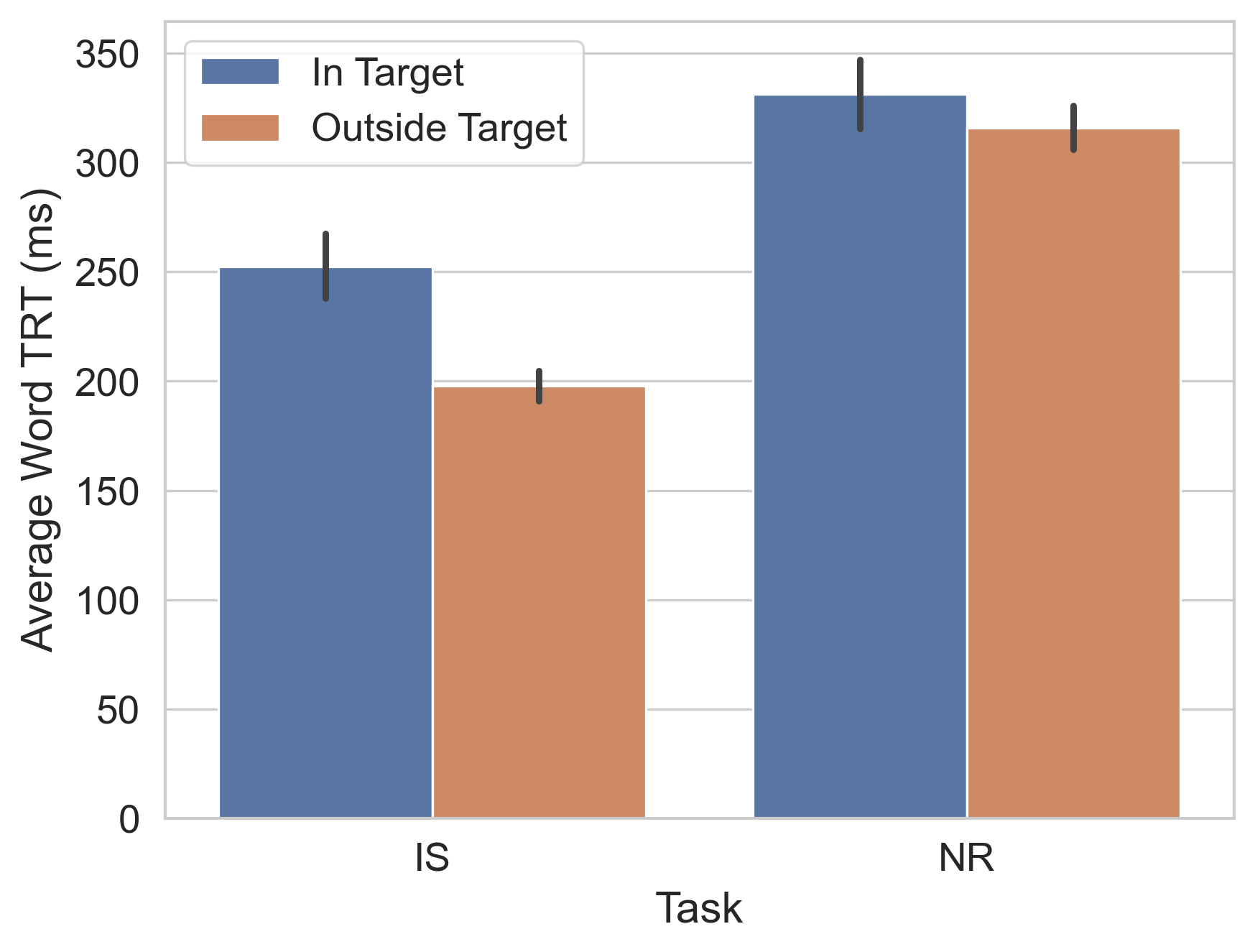}
    \captionof{figure}{Average TRT per word \textcolor{black}{for both information-seeking reading (IS) and normal reading (NR)}, for words within the target AOI and outside the target region. Lines on the bars represent the standard deviation. Values above the 99\% quantile are removed as outliers.}
    \label{fig:bar-plot-trt-is-vs-nr}
\end{figure}

\textcolor{black}{We further focus on the difference between the fixation duration within and outside of the target AOI by} analyzing the averaged TRT per word over all trials and grouping them by reading task (see Figure \ref{fig:bar-plot-trt-is-vs-nr}). Not only do participants spend less time on each word in the IS task over the NR task, but they spend slightly more time looking at words in the target region rather than on the rest of the text. This is in line with the results from \citet{malmaud-etal-2020-bridging}, whose study using a commercial eye-tracker in a lab setup yielded the same results. \textcolor{black}{Eye movements in \citet{malmaud-etal-2020-bridging} were recorded using an EyeLink 1000+ eye tracker (SR Research) at a sampling rate
of 1000Hz.} Furthermore, \citet{shubi2023eye} confirm that reading times and sensitivity to the linguistic properties of the text are strongly conditioned on the reading task.

\subsubsection{Correctness Classification}\label{sec:classifier}
With these findings in mind, we train a binary classifier to predict whether a correct answer has been given to a trial. We use the eye movement measures (1) - (6) as well as average TRT per word inside and outside the target \textcolor{black}{AOI} as input features. This results in a feature vector of $d=8$. For training the classifier, trials without any fixations on the text are removed ($\approx1\%$ of total data, $n=24$). 
Additionally, we experiment with adding two structural text features to the feature vectors, namely, \textit{token count} and \textit{average token length}. In these experiments $d=10$.

We then test three different classifiers from \textit{sklearn}\footnote{\url{https://scikit-learn.org/stable/}}: (i) SVC, a support vector machine classifier, (ii) Logistic Regression and (iii) Random Forests. We also use a random baseline classifier, which outputs the labels in the training data ($0=incorrect\,answer$, $1=correct\,answer$) in a uniform manner.
The parameters for all classifiers are set to their defaults from \textit{sklearn}. We split the data by task, NR ($n=1763$, $0=771$, $1=992$) and IS ($n=1743$, $0=348$, $1=1395$), and use a shuffled train ($80\%$) and test split ($20\%$). As there are substantially more positive labels, especially for the IS task, we decide to balance the labels by artificially up-sampling negative examples from the training split to augment the data, resulting in a balanced training set.

We then train each of the classifiers on this augmented training set and predict the labels of the test set. We repeat this procedure $10$ times with different random seeds. 
We provide the results for two scenarios: Filtering the data only from ``approved'' participants, i.e., participants with \textcolor{black}{an answer accuracy of $\geq50\%$}, and data from all participants.
We report the averaged results in Tables \ref{tab:class-results-approved} and \ref{tab:class-results-all}, respectively.

For the NR task, when filtering for ``approved'' participants, \textcolor{black}{the classifiers perform similarly to the baseline when using only eye movement features}, but the results improve when adding the two text features. The Random Forest classifier yields the highest result, an F1-score of \textcolor{black}{0.62}. \textcolor{black}{The other classifiers also show improvements when the text features are included, but remain below the Random Forest}. When including all participants, the results are \textcolor{black}{similar, however, all classifiers perform slightly worse}. Thus, the results indicate that the features used are insufficient to discriminate whether a participant answers correctly to the question in the given trial. This can result from the quality of the data, the selection of the features, or the nature of the NR task.

For the IS task, when filtering for ``approved'' participants, the Random Forest classifier yields an F1-score of 0.68 using eye movement features only and 0.74 when including both eye movement and text features. This is a substantial improvement over the baseline and also an improvement over the other classifiers.
When including all participants, the F1-scores are lower, but again we note the positive impact of the text features and the superiority of the Random Forest classifier. These results show that when filtering for participants with a high \textcolor{black}{answer accuracy}, the data is more biased toward this class.

To sum up, in the information-seeking task, a classifier can distinguish accurately whether a correct or incorrect answer was given to the question of a specific trial. Moreover, the Random Forest classifiers yield the best results, and adding structural text features is useful in all cases. Results for each language individually can be found in Appendix \ref{app:classification}.
Therefore, returning to our hypothesis, the results indicate that while IS reading provides reasonable proxies for human rationales, this does not seem to be the case for the NR task, where only the inclusion of text features can help classifiers perform better than the baseline. Hence, task-based reading is more promising to automatically extract human rationales from eye movement information. Possibly, the best strategy is to align the NLP task for which a model is to be trained with the reading task as we do here by linking information-seeking reading with question-answering.

\begin{table}[t]
\centering
\caption{Accuracy and weighted F1-scores for classifier models trained to predict whether a correct answer was given for the corresponding text in the IS and NR tasks \textbf{from ``approved'' participants ($n=353$)}. Results are averaged across 10 runs (standard deviation $\sigma\le0.04$ for all NR models and $\sigma\le0.07$ for the IS models).}
\label{tab:class-results-approved}
\begin{tabular}{l|llll|llll}
\toprule
\small
\textbf{Model} & \multicolumn{4}{c}{\textbf{NR}} & \multicolumn{4}{c}{\textbf{IS}} \\
 & \multicolumn{2}{c}{gaze} & \multicolumn{2}{l}{gaze+text} & \multicolumn{2}{c}{gaze} & \multicolumn{2}{c}{gaze+text} \\
 & Acc & F1 & Acc & F1 & Acc & F1 & Acc & F1 \\\midrule
Random       &  0.51  & 0.51  &  0.50  & 0.50  & 0.51  & 0.56  &  0.49  & 0.54   \\\midrule
%Majority     &  \textbf{0.56}  & 0.40  &  0.56  & 0.40  & \textbf{0.80}  & \textbf{0.71}  &  \textbf{0.80}  & 0.71   \\
SVM          &  0.55  & 0.53  &  0.59  & 0.59  & 0.64  & 0.67  &  0.63  & 0.66   \\
Log. Reg.    &  0.54  & \textbf{0.54}  &  0.57  & 0.64  & 0.67  & 0.64  &  0.58  & 0.63  \\
Rand. Forest &  0.53  & 0.53 &  \textbf{0.62} &\textbf{0.61}  & \textbf{0.69}  & \textbf{0.68}  &  \textbf{0.74}  & \textbf{0.74}  \\\bottomrule
\end{tabular}
\end{table}

\begin{table}[t]
\centering
\caption{Accuracy and weighted F1-scores for classifier models trained to predict whether a correct answer was given for the corresponding text in the IS and NR tasks \textbf{including all participants ($n=487$)}. Results are averaged across 10 runs (standard deviation $\sigma\le0.04$ for all models).}
\label{tab:class-results-all}
\begin{tabular}{l|llll|llll}
\toprule
\textbf{Model} & \multicolumn{4}{c}{\textbf{NR}} & \multicolumn{4}{c}{\textbf{IS}} \\
 & \multicolumn{2}{c}{gaze} & \multicolumn{2}{l}{gaze+text} & \multicolumn{2}{c}{gaze} & \multicolumn{2}{c}{gaze+text} \\
 & Acc & F1 & Acc & F1 & Acc & F1 & Acc & F1 \\\midrule
Random       &  0.50  & 0.50  &  0.51  & 0.51  &  0.50 & 0.51  &  0.51  & 0.52  \\\midrule
%Majority     &  0.53  & 0.37  &  0.53 & 0.37  &  \textbf{0.65} & 0.52  &  \textbf{0.65}  & 0.52  \\
SVM      &  \textbf{0.54}  & 0.44  &  \textbf{0.61} & 0.57  &  \textbf{0.60} & \textbf{0.59} &  0.60  & 0.61  \\
Log. Reg.    &  0.53  & 0.41  &  0.61  & 0.60  &  0.58  & 0.58  &  0.57  & 0.58  \\
Rand. Forest &  0.52  & \textbf{0.51} &  \textbf{0.61}  & \textbf{0.61}  &  0.56  & 0.57  &  \textbf{0.65}  & \textbf{0.65}  \\\bottomrule
\end{tabular}
\end{table}

\section{Discussion}

We compile and share a new webcam eye-tracking dataset including reading data from 600 participants in four languages. The results obtained in the \textsc{WebQAmGaze} dataset reveal similar cognitive processing information to what was previously considered to be only possible with a commercial eye-tracker in a lab environment, \textcolor{black}{including higher fixation duration on longer words and differences in the total reading times between the two reading tasks}. We show promising results when comparing our data to that of a high-quality setup, showing that word-level fixation duration measures of webcam data correlate to those collected with a commercial eye-tracker.
Furthermore, we find that eye movement measures from task-specific reading align with the relevant text spans for question-answering. 

\textcolor{black}{In this section, we highlight the challenges encountered with \textsc{WebQAmGaze} with respect to the methodology (Section \ref{sec:method-chall}) and the data quality (Section \ref{sec:quality-chall}), as well as potential directions for future work (Section \ref{sec:future-work}). }

\subsection{Methodological Challenges}\label{sec:method-chall}

With the proposed \textsc{WebQAmGaze} data collection process, we describe a solution to present the text stimuli in a manner that maintains consistency across the diversity of possible computer setups. However, other factors are more difficult to control. First, \textit{WebGazer} requires a combination of a webcam with a good sampling rate ($\approx30$Hz) and a computer with a sufficient CPU and RAM, which is hard to control in the current setting. We attempt to counteract this challenge by using the reported sampling rate, which is a good indicator of how well \textit{WebGazer} is functioning. 

In terms of experiment reporting, we follow the recommended empirical guidelines for reporting eye-tracking experiments wherever possible \cite{holmqvist2023eye,dunn2023minimal}. However, we note that for webcam-based gaze experiments the guidelines must be adapted to include a range of values for categories such as screen sizes and sampling rates (rather than one specific value for a single tracking device). Accordingly, the information collected about screen resolutions allowed us to evaluate the inaccuracy in terms of pixels, but not in absolute distances. In future work, a guideline specific to webcam-based eye-tracking experiments is desirable.

Furthermore, we face the issue that while we perform multiple calibration steps throughout the experiment to ensure that we correct for possible head movements by participants, we are unable to validate the accuracy of the eye-tracking throughout the experiment, as this would add a considerable amount of time to the total experiment and could result in possible loss of interest or focus from the participants. An intermediate solution could be to add a validation step at the end of the experiment to check how much the accuracy degraded over time.
%It is also worth mentioning that \textit{WebGazer} itself might crash or run into issues while the experiment is ongoing. This can result in data loss for certain trials or an uncompleted experiment.

Finally, and most importantly, it is difficult to know how engaged the participants are during the experiment. Nothing stops participants from getting distracted, and it is difficult to assess how closely they follow the instructions given. The minimum number of correct answers required for payment is the only controlling factor we implemented for the \textit{mTurk} participants. Nevertheless, observing our data analysis results, we can see that following the proposed filtering steps yields cleaner data.

\subsection{Data Quality Challenges}\label{sec:quality-chall}

The texts and questions we use in \textsc{WebQAmGaze} come from existing NLP datasets, which have their own limitations. Namely, XQuAD \citep{artetxe-etal-2020-cross} contains open-field questions that are sometimes formulated in an unclear manner. As the texts are originally collected from Wikipedia, they can contain typos or other text marks, which can cause distraction or confusion during reading. The annotations of the target spans can be ambiguous or only partially correct, which sets an upper bound for human rationale prediction and for the correctness classification presented in Section \ref{sec:classifier}. On the contrary, since the texts represent naturally occurring language from the web, it increases the ecological validity of the experiment and brings it closer to an everyday reading scenario.
The MECO corpus \citep{siegelman_expanding_2022_meco} contains true/false comprehension questions, which in some cases can be answered by relying on commonsense or world knowledge rather than by carefully reading the text.

\begin{figure*}[t]
     \centering
     \begin{subfigure}[b]{0.43\textwidth}
         \centering
         \includegraphics[width=\textwidth]{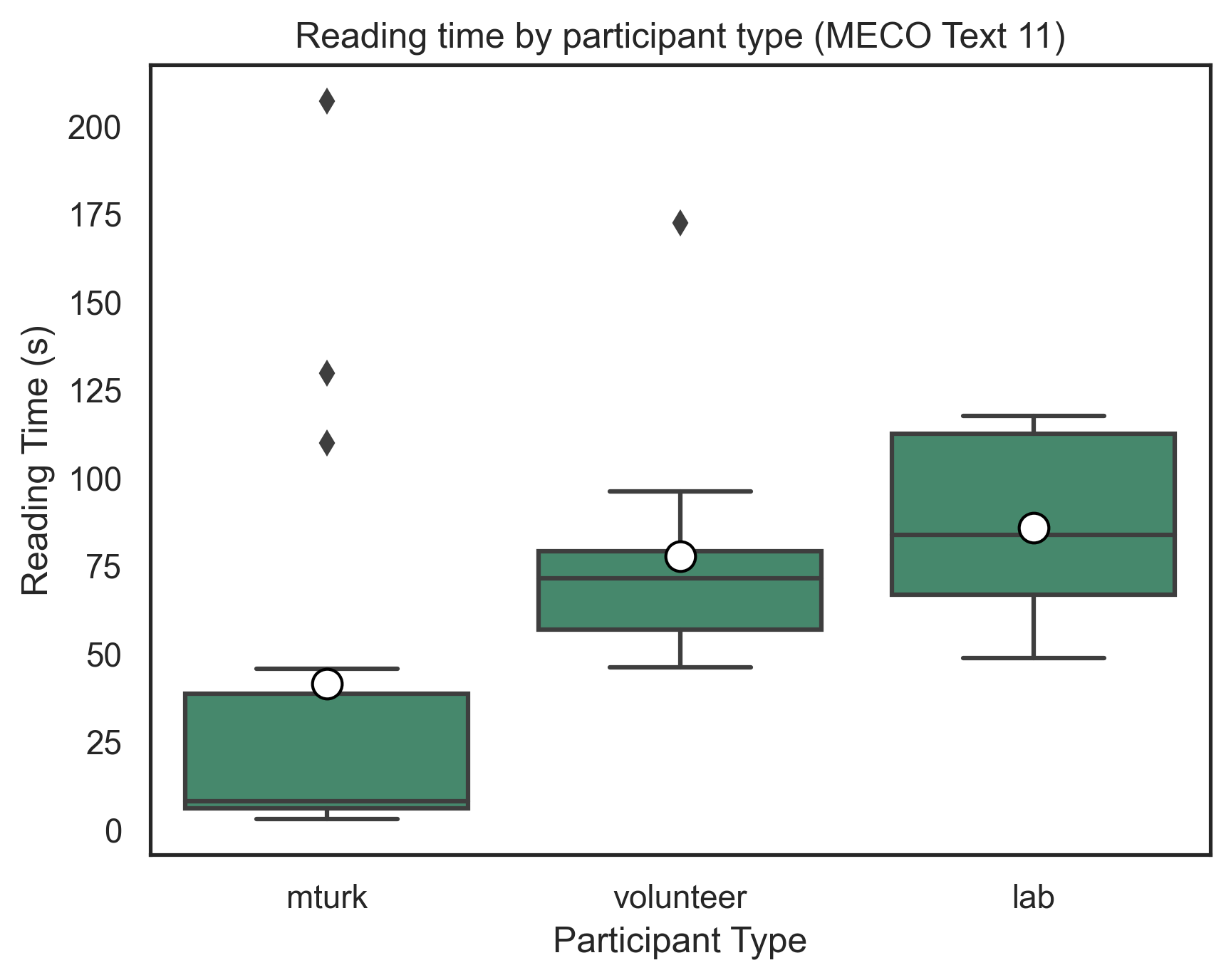}
         \caption{Reading time on the English MECO text 11 across participant groups, including 15 \textit{mTurk} participants,
         %from set (EN) 14 and 18 ($n=15$),
         10 online volunteers, and 5 lab volunteers.}
         \label{fig:time-reading-text11}
     \end{subfigure}
     \hfill
     \begin{subfigure}[b]{0.43\textwidth}
         \centering
         \includegraphics[width=\textwidth]{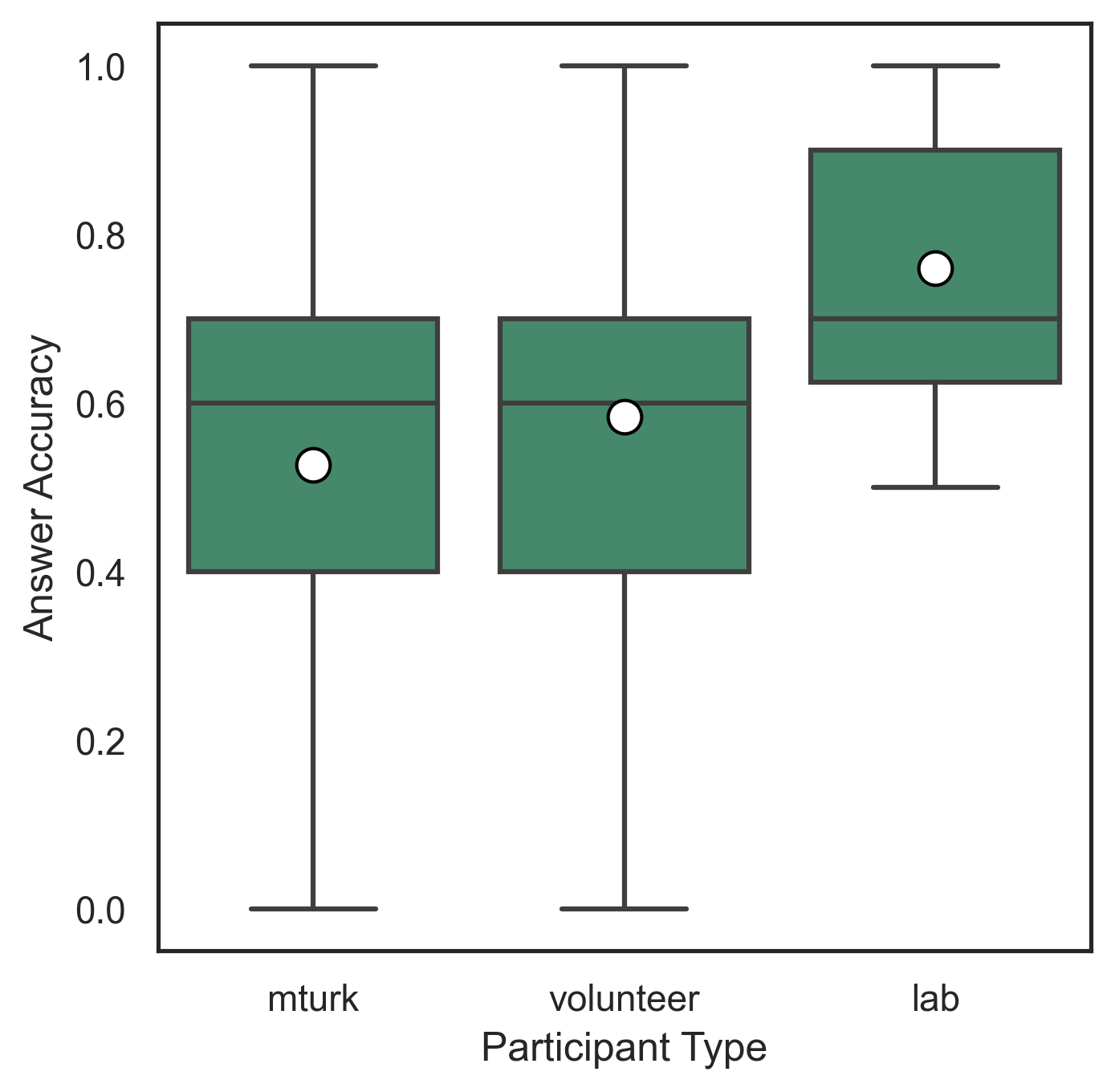}
         \caption{Answer accuracy across participant groups in the whole \textsc{WebQAmGaze} dataset ($n=600$), 350 \textit{mTurk}, 240 online volunteers, and 10 lab volunteers.}
         \label{fig:correctness-plot}
     \end{subfigure}
     \hfill
        \caption{\textcolor{black}{Comparison of the participant groups within the \textsc{WebQAmGaze} dataset.}}
        \label{fig:dif-participants}
    \end{figure*}

\textcolor{black}{Another consideration for the \textsc{WebQAmGaze} data collection is the fact that participants on \textit{Amazon Mechanical Turk} were paid to participate in the experiment only if they achieved $\geq5$ correct answers. The three participant populations described in Section \ref{sec:crowd-sourcing} allow us to analyze the impact of monetary reimbursement in terms of reading time and number of correct answers. We plot the reading time for one of the MECO texts in Figure \ref{fig:time-reading-text11}. The MECO texts are the longest in the dataset, and we see a considerable difference between the reimbursed \textit{mTurk} participants and the volunteers, showing that \textit{mTurk} participants on average spend less time reading the text than the other populations, which can be explained by the incentive of completing the task as quick as possible to optimize the monetary reward. Additionally, Figure \ref{fig:correctness-plot} shows that the lab-controlled volunteers achieved the highest answer accuracy, possibly due to the more supervised experiment setup eliciting higher focus from the participants.}

\textcolor{black}{To conclude, while it is less expensive to collect more data in a crowd-sourcing setup, we also note that there is a large number of participants and gaze data that is removed due to the quality criteria we apply during data processing. However, as one of the first attempts at collecting such a dataset, our contribution provides valuable initial evidence of the type of challenges that arise in webcam-based reading studies. We point to possible approaches to addressing these in the following section.} 

\subsection{Future Work} \label{sec:future-work}

A potential path for improvement is the inclusion of superior algorithms for fixation detection, for instance, machine-learning-based event detection methods \citep{birawo2022review,nuraini2021machine}. \textcolor{black}{With our simple approach, we find similar patterns to those observed when compared to high-quality eye-tracking data, e.g., a clear effect in word length}. However, it can be improved further with more complex methods, such as correcting the gaze points by taking line height, text layout, distribution of fixations, or accuracy and sampling rate of \textit{WebGazer} into account when performing the fixation cleaning and merging. 

\textcolor{black}{A preliminary survey could be used to gain information about the participant's webcam setup as well as to improve the data quality by performing a calibration step to calculate the \textit{WebGazer} calibration accuracy and sampling rate before performing the entire experiment. This would allow pre-filtering cases where the participant's webcam setup is of lower quality.} 

Finally, as described by \citet{malmaud-etal-2020-bridging}, the extracted eye movement measures
can be used in combination with computational language models, such as BERT \citep{devlin-etal-2019-bert}, to investigate if this leads to more human-like reasoning in the model.
Given that our data is collected on texts that contain annotated rationales, it is possible to analyze if \textsc{WebQAmGaze} improves the performance and explainability of these models in a question-answering application. 

\section{Conclusion} 

We present a novel approach to collecting low-cost eye-tracking data from webcam recordings. We compile the \textsc{WebQAmGaze} dataset, which includes word-level eye movement measures from reading tasks. We demonstrate that the data reflects linguistic patterns that have been corroborated by previous studies.
%namely, in our comparison with high-quality eye-tracking recordings. 
In line with webcam eye-tracking research performed on other types of visual tasks \citep{wisiecka2022comparison}, we also find that while error measurements are higher, the data follows the theoretical expectations.  

We show that webcam eye-tracking can be used to predict the correctness of participants' responses in a task-specific context, paving the way to a more efficient collection of human rationales for explainable AI. Knowing where readers look can help to explain machine behavior in terms of human cognitive processes \citep{ikhwantri2023looking,hollenstein-beinborn-2021-relative}. 

Lastly, the online crowd-sourcing approach is useful for collecting data from a wider population range and is advantageous in terms of ease of access, both physical and related to hardware. While webcam eye-tracking still lags behind with respect to data quality %and makes it difficult to analyze millisecond-accurate reading behavior,
, it does reflect similar patterns as high-quality data and is therefore a viable alternative, especially when leveraging eye movement data for evaluating and improving computational language processing models.

\section*{Acknowledgments}

This research was conducted with the support of a grant from the Carlsberg Foundation.
For their help in annotating the MECO spans, we thank Marina Pinzhakova, Ece Takmaz, and Eleni Ioakeim. We further appreciate the help in distributing the experiment from Ana Matić Škorić and other members of the MultiplEYE COST Action.

\section*{Ethics approval} 
The experiment design for this data collection was approved by the Ethics Commission of the Faculty of Humanities of the University of Copenhagen.
\section*{Consent to participate} 
All participants gave consent for their participation.
\section*{Consent for publication} 
All participants gave consent for the publication and use of their data for research purposes.
\section*{Availability of data and code} 
The data and code for all experiments are openly available: \\
\url{https://github.com/tfnribeiro/WebQAmGaze}

%\noindent
%If any of the sections are not relevant to your manuscript, please include the heading and write `Not applicable' for that section. 


\bibliography{custom}

\pagebreak
\begin{appendices}

\section{Experiment Instructions}\label{app:instructions}

\begin{figure}[h]
    \centering
\includegraphics[width=0.8\textwidth]{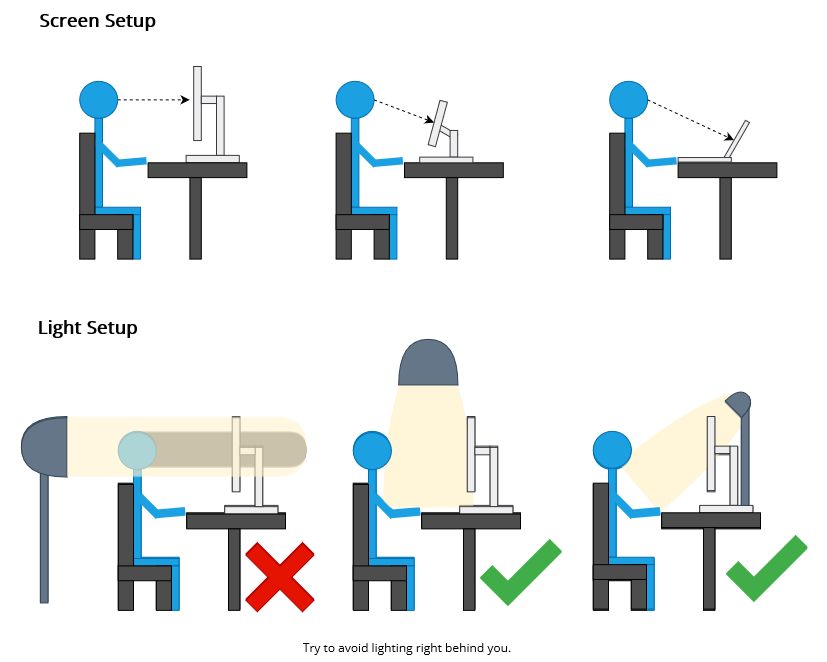}
    \caption{Instructions for sitting position.}
    \label{fig:sitting-instructions}
\end{figure}

\begin{figure}[h]
    \centering
\includegraphics[width=0.8\textwidth]{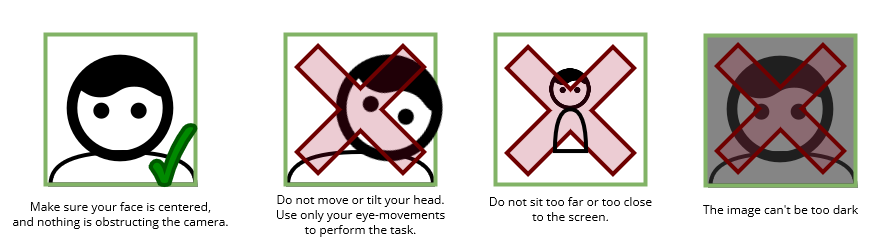}
    \caption{Instructions for the positioning of the head.}
    \label{fig:head-instructions}
\end{figure}

Figures \ref{fig:sitting-instructions} and \ref{fig:head-instructions} present the graphics used in the experimental setup to describe the optimal sitting and head position to the participants. These instruction graphics were inspired by \citet{schneegans2021exploring}.

\clearpage
\section{Calibration grids}\label{app:calibration-grid}

\begin{figure}[!h]
    \centering
    \includegraphics[width=0.75\textwidth]{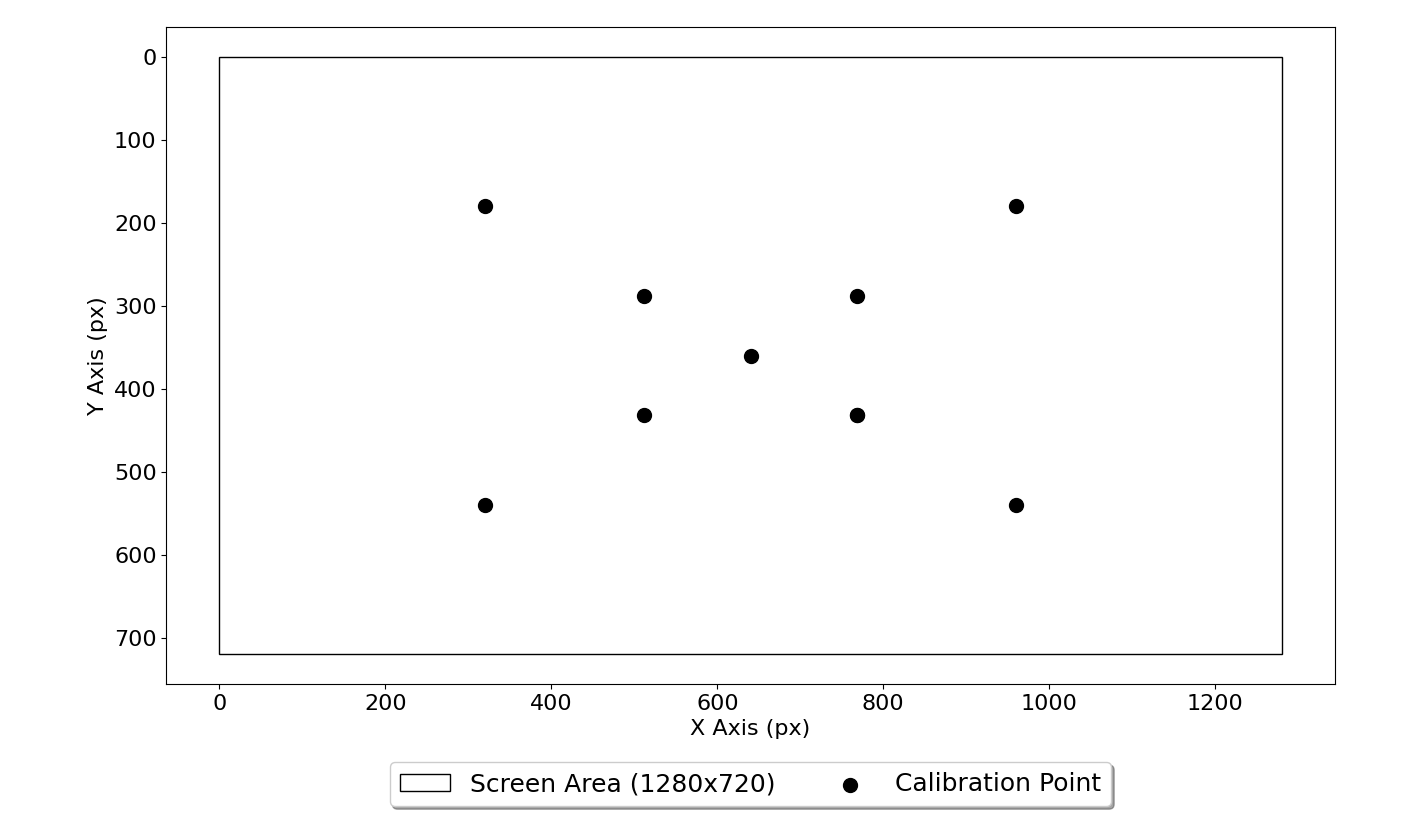}
    \caption{9-point calibration grid used for a participant with a screen size of 1280x720, at the start of the experiment. The grid scales so that the points correspond to the same percentage on larger screens.}
    \label{fig:calibration-grid}
\end{figure}

\begin{figure}[!h]
    \centering
    \includegraphics[width=0.75\textwidth]{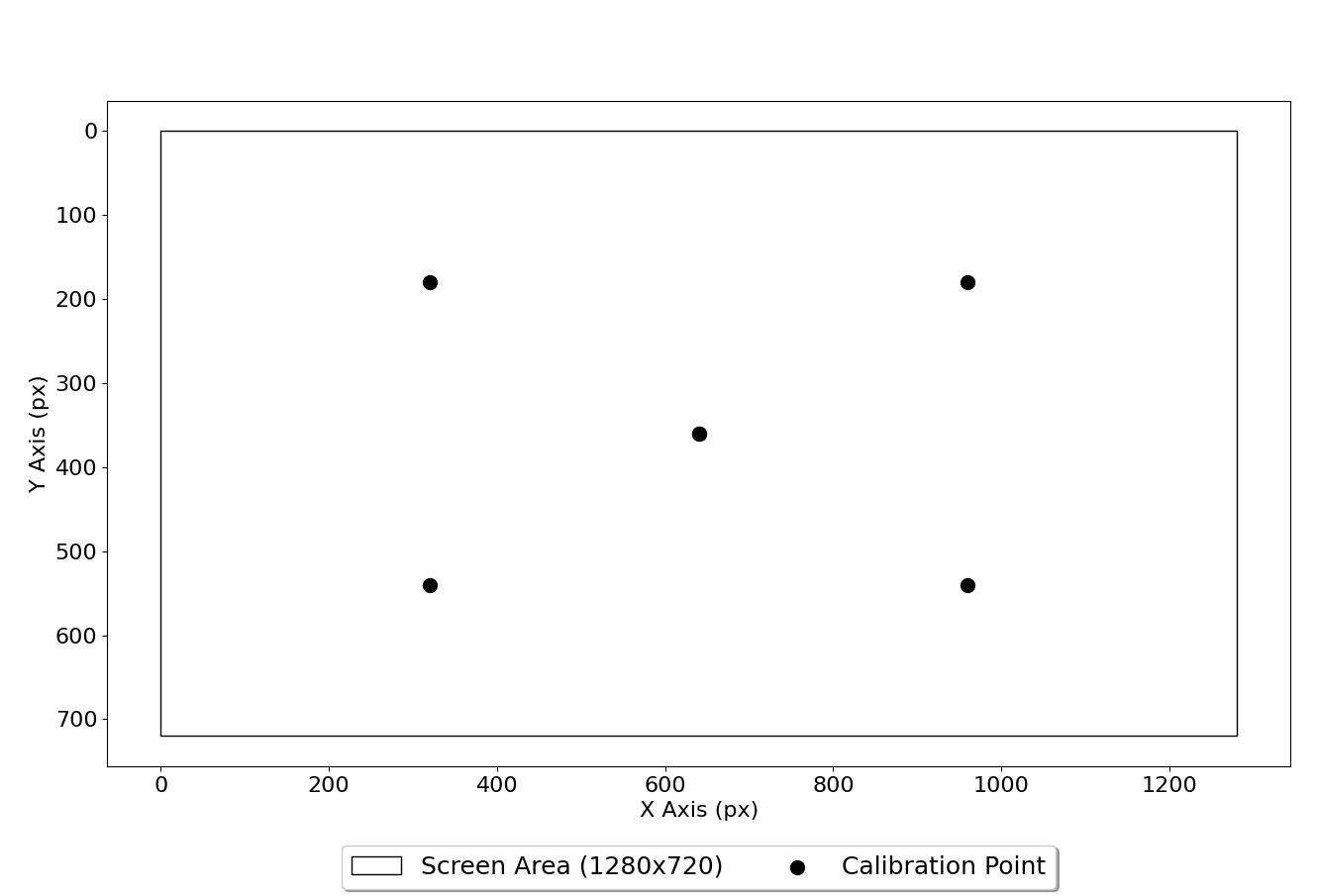}
    \caption{5-point quick calibration grid used for a participant with a screen size of 1280x720, during the experiment. The grid scales so that the points correspond to the same percentage on larger screens.}
    \label{fig:calibration-grid-quick}
\end{figure}

Figures \ref{fig:calibration-grid} and \ref{fig:calibration-grid-quick} show the calibration grid used at the start and during quick calibration, respectively.
% Maybe we should say that we opt to have the calibration points more closely to the center as our ROI will always be centered on the screen? 

\newpage
\section{Validation Results}\label{app:validation}

\begin{figure}[!h]
    \centering
    \includegraphics[width=\textwidth]{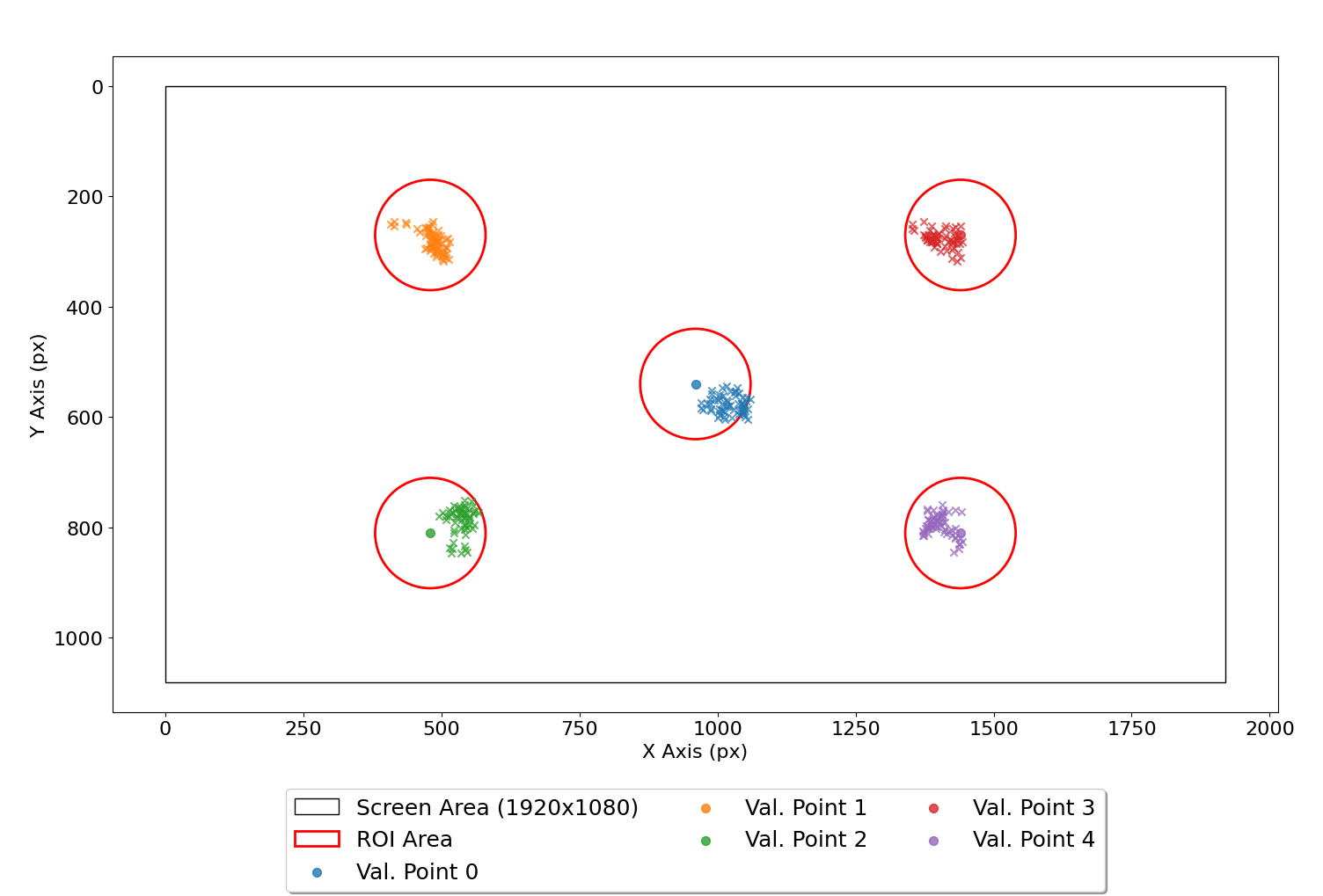}
    \caption{Validation plot for the participant with ID A2N9U74YIPDQ9F. The calibration accuracy is 99.7\%, and the average offset is 29.6 pixels with a screen resolution of 1920x1080.}
    \label{fig:good-validation}
\end{figure}

\begin{figure}[!h]
    \centering
    \includegraphics[width=\textwidth]{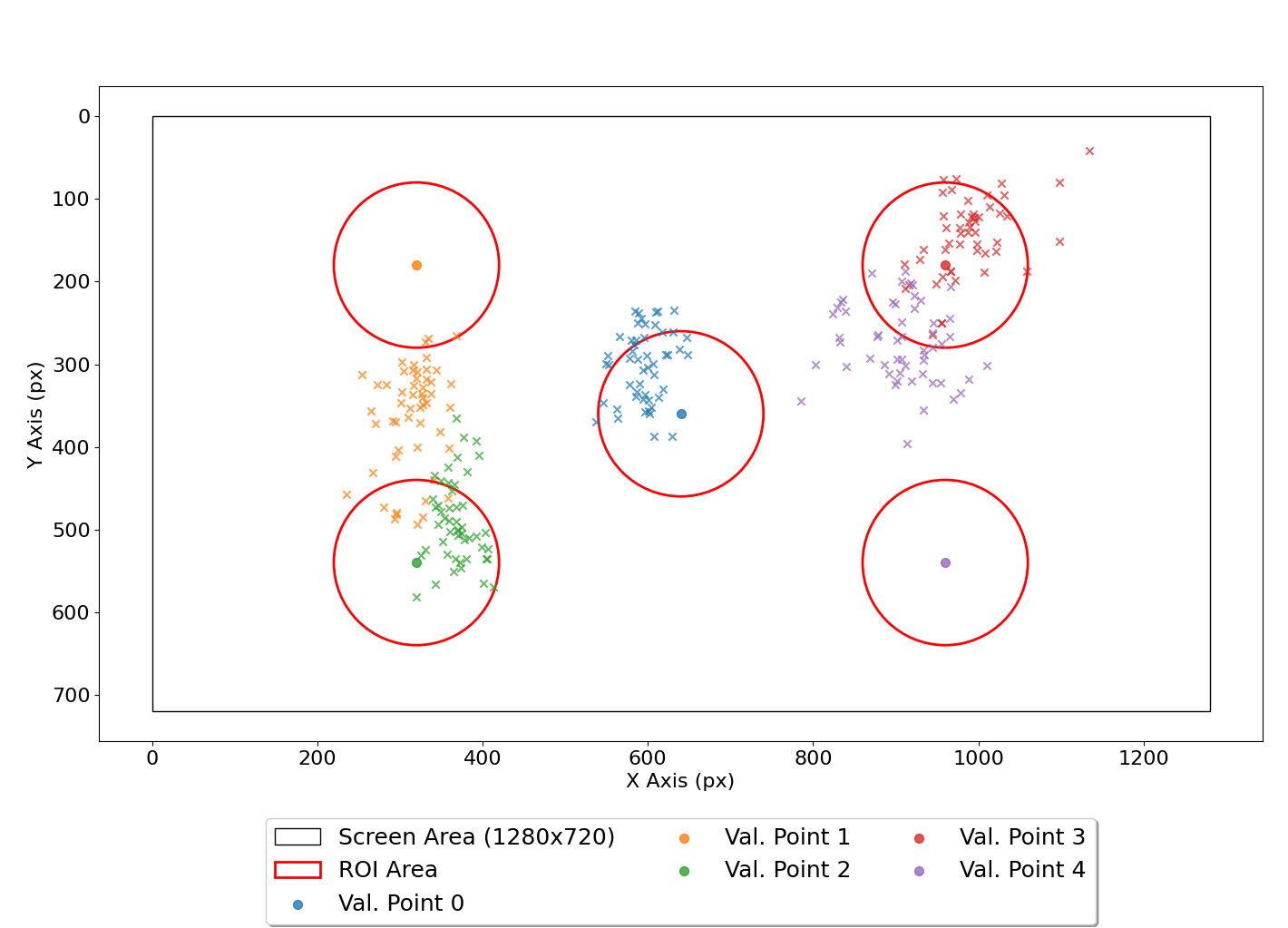}
    \caption{Validation plot for the participant with ID A39SK1E6IMQBD5. The calibration accuracy is 43.3\%, and the average offset is 59.1 pixels with a screen resolution of 1280x720.}
    \label{fig:bad-validation}
\end{figure}

Figures \ref{fig:good-validation} and \ref{fig:bad-validation} show examples of good and bad validation results.

\clearpage
\section{Significance Testing}\label{app:ttests}

Tables \ref{tab:t-tests-EN}, \ref{tab:t-tests-DE}, \ref{tab:t-tests-ES}, \ref{tab:t-tests-TR} present the t-test results for each individual language, English, German, Spanish, and Turkish, respectively. p-values are obtained for the independent t-tests performed by separating the groups into correct or incorrect responses given to a trial, aggregated over the NR or IS tasks. Significant results ($p<0.05$) are marked in bold.
% EN: significant results only for IS, 1 more feature significant with filter
% ES: significant results only for IS, same with filter
% DE: no significant results, 
% TR: NR feature 3 significant, 3 fts for IS, very small sample size, same with filter

\begin{table}[h]
\centering
\caption{Significance testing for \textbf{English}. The number of trials is 
%$n=1540$ (930 with 20\% filter) 
$n=770$ or $n=465$ for each task, without and with the 20\% \textcolor{black}{calibration} accuracy filter, respectively.
%for feature (6) the number is $n=1450 (880)$, due to some lack of TRT text for some participants.
}
\label{tab:t-tests-EN}
\begin{tabular}{l|rr|rr}
\toprule
\textbf{Feature} & \multicolumn{2}{c|}{\textbf{NR}} & \multicolumn{2}{c}{\textbf{IS}} \\
 & all & 20\% filter & all & 20\% filter \\\midrule
(1) Fixation on target AOI             & 0.692 & 0.407 & 0.053 & \textbf{0.048}  \\
(2) Number of fixations                & 0.905 & 0.452 & \textbf{0.010} & $\mathbf{<0.001}$   \\
(3) Target fixation ratio    & 0.817 & 0.789 & 0.152 & 0.121  \\
(4) TRT on paragraph AOI                    & 0.721 & 0.439 & \textbf{0.010} & $\mathbf{<0.001}$   \\
(5) TRT on target AOI                  & 0.492 & 0.427 & 0.061 & \textbf{0.043}  \\
(6) Target TRT ratio            & 0.754 & 0.599 & 0.132 & 0.144  \\
(7) Total trial time               & 0.149 & 0.083 & \textbf{0.002} & \textbf{0.004}  \\\bottomrule
\end{tabular}
\end{table}

\begin{table}[h]
\centering
\caption{Significance testing for \textbf{Spanish}. The number of trials is 
%$n=570 (20\% n=410)$ 
$n=285$ or $n=205$ for each task, without and with the 20\% \textcolor{black}{calibration} accuracy filter, respectively. 
%for feature (6) the number is $n=550$, due to some lack of TRT text for some participants. 
}
\label{tab:t-tests-ES}
\begin{tabular}{l|rr|rr}
\toprule
\textbf{Feature} & \multicolumn{2}{c|}{\textbf{NR}} & \multicolumn{2}{c}{\textbf{IS}} \\
 & all & 20\% filter & all & 20\% filter \\\midrule
(1) Fixation on target AOI             & 0.366 & 0.571 & 0.307 & 0.244  \\
(2) Number of fixations              & 0.086 & 0.241 & \textbf{0.005} & \textbf{0.004}  \\
(3) Target fixation ratio     & 0.450 & 0.393 & 0.591 & 0.667  \\
(4) TRT on paragraph AOI                    & 0.340 & 0.411 & \textbf{0.011} & \textbf{0.004}  \\
(5) TRT on target AOI                  & 0.977 & 0.743 & 0.332 & 0.159  \\
(6) Target TRT ratio             & 0.742 & 0.438 & 0.482 & 0.559  \\
(7) Total trial time               & 0.073 & 0.098 & $\mathbf{<0.001}$ & $\mathbf{<0.001}$  \\\bottomrule
\end{tabular}
\end{table}

\begin{table}[h]
\centering
\caption{Significance testing for \textbf{German}. The number of trials is %$n=210 (160)$ 
$n=105$ or $n=80$ for each task, without and with the 20\% \textcolor{black}{calibration} accuracy filter, respectively. 
%for feature (6) the number is $n=200 (150)$, due to some lack of TRT text for some participants. In bold, we highlight the significant results ($p<0.05$).
}
\label{tab:t-tests-DE}
\begin{tabular}{l|rr|rr}
\toprule
\textbf{Feature} & \multicolumn{2}{c|}{\textbf{NR}} & \multicolumn{2}{c}{\textbf{IS}} \\
 & all & 20\% filter & all & 20\% filter \\\midrule
(1) Fixation on target AOI             & 0.645 & 0.709 & 0.789 & 0.504  \\
(2) Number of fixations                   & 0.966 & 0.879 & 0.418 & 0.726  \\
(3) Target fixation ratio      & 0.639 & 0.848 & 0.962 & 0.949  \\
(4) TRT on paragraph AOI                    & 0.846 & 0.976 & 0.386 & 0.761  \\
(5) TRT on target AOI                 & 0.724 & 0.884 & 0.850 & 0.734  \\
(6) Target TRT ratio             & 0.671 & 0.760 & 0.990 & 0.989  \\
(7) Total trial time               & 0.714 & 0.762 & 0.671 & 0.805  \\\bottomrule
\end{tabular}
\end{table}

\begin{table}[h]
\centering
\caption{Significance testing for \textbf{Turkish}. The number of trials is %$n=1210 (620)$ 
$n=605$ or $n=310$ for each task, without and with the 20\% \textcolor{black}{calibration} accuracy filter, respectively.  
%for feature (6) the number is $n=1170 (620)$, due to some lack of TRT text for some participants. 
}
\label{tab:t-tests-TR}
\begin{tabular}{l|rr|rr}
\toprule
\textbf{Feature} & \multicolumn{2}{c|}{\textbf{NR}} & \multicolumn{2}{c}{\textbf{IS}} \\
 & all & 20\% filter & all & 20\% filter \\\midrule
(1) Fixation on target AOI               & 0.328 & \textbf{0.047} & \textbf{0.021} & \textbf{0.026}  \\
(2) Number of fixations               & 0.351 & 0.622 & \textbf{0.008} & $\mathbf{<0.001}$  \\
(3) Target fixation ratio     & \textbf{0.001} & \textbf{<0.001} & 0.525 & 0.203  \\
(4) TRT on paragraph AOI                    & 0.625 & 0.705 & \textbf{0.011} & $\mathbf{<0.001}$  \\
(5) TRT on target AOI                  & 0.285 & 0.094 & \textbf{0.021} & \textbf{0.023}  \\
(6) Target TRT ratio            & \textbf{0.002} & \textbf{0.001} & 0.306 & 0.208  \\
(7) Total trial time               & 0.340 & 0.542 & $\mathbf{<0.001}$ & $\mathbf{<0.001}$  \\\bottomrule
\end{tabular}
\end{table}

\clearpage
\section{Classification Results}\label{app:classification}

Table \ref{tab:class-results-languages} presents the results for the random baseline and the forest classifier models for \textbf{all languages individually}, trained to predict whether a correct answer was given for the corresponding sample in the IS and NR tasks.

\begin{table}[h]
\centering
\caption{Accuracy and weighted F1-scores for the random baseline and the random forest classifier models for \textbf{all languages individually}, trained to predict whether a correct answer was given for the corresponding sample in the IS and NR tasks \textbf{including all participants}. Results are averaged across 10 runs (standard deviation $\sigma\le0.04$ for all EN models, $\sigma\le0.11$ for DE, $\sigma\le0.07$ for ES, and $\sigma\le0.05$ for TR).}
\label{tab:class-results-languages}
\begin{tabular}{l|c|llll|llll}
\toprule
& & \multicolumn{4}{c}{\textbf{NR}} & \multicolumn{4}{c}{\textbf{IS}} \\
& & \multicolumn{2}{c}{gaze} & \multicolumn{2}{l}{gaze+text} & \multicolumn{2}{c}{gaze} & \multicolumn{2}{c}{gaze+text} \\
\textbf{Model} & \textbf{Participants} & Acc & F1 & Acc & F1 & Acc & F1 & Acc & F1 \\\midrule
Random   &    & 0.52 & 0.52 &  0.50  & 0.50  &  0.51  & 0.54  & 0.49  & 0.52   \\
English & 183& 0.53  & 0.53  &  0.64  & 0.61  &  0.60 & 0.60  & 0.63  & 0.62 \\\midrule
Random    &   & 0.53  & 0.53  & 0.51  & 0.51  &  0.48  & 0.48 & 0.50  & 0.51    \\
German & 45  & 0.56  & 0.51  & 0.70  & 0.68  &  0.51  & 0.50  & 0.61 & 0.61  \\\midrule
Random  &     & 0.51  & 0.51 & 0.52  & 0.52  &  0.48  & 0.49  & 0.50 & 0.51   \\
Spanish & 85 & 0.56  & 0.55  & 0.60 & 0.59  &  0.56  & 0.56  & 0.64 & 0.64 \\\midrule
Random   &    & 0.52  & 0.53  & 0.51  & 0.51 &  0.51  & 0.52  & 0.51  & 0.52 \\
\textcolor{black}{Turkish} & 174 & 0.54 & 0.54 & 0.61 & 0.61 & 0.59 & 0.59 & 0.68 & 0.69\\\bottomrule
\end{tabular}
\end{table}




\end{appendices}


\end{document}